%% file: main.tex
\documentclass{article} 
\usepackage{iclr2027_conference,times}

\input{math_commands.tex}

\usepackage{hyperref}
\usepackage{url}
\usepackage{graphicx}
\usepackage{algorithm}
\usepackage{graphicx}
\usepackage{booktabs}
\usepackage{capt-of}
\usepackage{algpseudocode}
\usepackage{tabularx}
\usepackage{booktabs}
\usepackage{tikz}
\usepackage{placeins}
\usepackage{arydshln}
\usepackage{tikz}
\usetikzlibrary{tikzmark}
\usetikzlibrary{arrows.meta,positioning}

\title{AutoLoCo: Communication Efficient Distributed LLM Training via Adaptive Synchronization} 

\author{%
Pengyu He$^{1,*}$
\quad Yan Zhang$^{2,*,\dagger}$
\quad Ruien Li$^{3}$
\quad Guangwen Yang$^{2}$ \\
$^{1}$Department of Electrical Engineering, California Institute of Technology \\
$^{2}$Department of Computer Science and Technology, Tsinghua University \\
$^{3}$Department of Computer Sciences, University of Wisconsin - Madison
}

\iclrfinalcopy 
\begin{document}

\maketitle

\begingroup
\renewcommand{\thefootnote}{\fnsymbol{footnote}}
\footnotetext[1]{Equal contribution.}
\footnotetext[2]{Correspondence to: Yan Zhang
\textless\texttt{yanzhang24@mails.tsinghua.edu.cn}\textgreater.}
\endgroup

\begin{abstract}
The pre-training of Large Language Models (LLMs) is increasingly conducted across multiple data centers. As training scales to a larger number of accelerators, the fraction of time spent on computation decreases, while the fraction spent on communication increases. Therefore, frequent synchronization becomes a growing
bottleneck. Local update methods reduce this cost by allowing workers
to perform several optimizer steps between synchronizations. Most local update methods set the number of local optimizer steps between synchronizations before training and keep this interval fixed throughout the run. However, the best interval can change during the entire train process. If the interval and optimizer are adapted to the current training state, the communication frequency is reduced while maintaining the training performance. In this work, we introduce AutoLoCo, an adaptive training framework to reduce communication in LLM training. It adapts the local interval using scalar training statistics and corrects each outer update. Our method is motivated by two observations: 1) the appropriate local interval varies across training stages, and 2) changing the number of inner steps per interval creates a mismatch with an unchanged outer optimizer, requiring a correction to the outer update. We optimize this mismatch by correction of the outer optimizer for the momentum and the learning rate using the accumulated inner learning rate. Our experiments under communication constraints demonstrate
that AutoLoCo reduces communication frequency by 27\% relative to
DiLoCo while maintaining training performance.
\end{abstract}

\section{Introduction}
\label{sec:introduction}

Developing high-capability Large Language Models (LLMs) requires substantial data and
computation, making training efficiency a central concern
\citep{dubey2024llama3}. LLMs' training increasingly relies on distributed accelerators,
while useful compute may be spread across datacenters, institutions, and
regions. These resources cannot always communicate as one tightly
integrated cluster. Each synchronization incurs blocking latency and transfers updates for the entire model over links with limited bandwidth.
Communication becomes a central constraint when training spans geographically
distributed resources
\citep{douillard2023diloco,jaghouar2024opendiloco}.

Existing methods reduce either the amount or the frequency of communication.
DiLoCo follows the second approach. Each worker performs \(H\) local AdamW
steps before a Nesterov outer optimizer updates the shared model from averaged
pseudo-gradients \citep{douillard2023diloco}. By synchronizing only periodically, DiLoCo enables training across
loosely connected compute resources without requiring frequent
model-wide communication. However, standard DiLoCo fixes the communication interval throughout
training. A short interval can incur unnecessary synchronization, when a long
interval can allow excessive disagreement between local updates. Adaptive Periodic Averaging, STL-SGD, and the Quadratic Synchronization
Rule adapt synchronization periods using worker disagreement or
learning-rate schedules
\citep{jiang2020adaptive,shen2021stlsgd,gu2024quadratic}.
Related work also highlights interactions among the local interval,
outer momentum, and outer learning rate \citep{kallusky2025snoo}.
The remaining design question is how to coordinate online interval selection with outer updates adapted to each interval in DiLoCo.

The main challenge is to reduce synchronization while preserving training
quality. As the inner learning rate decays and individual updates become
smaller, later training stages tolerate longer local intervals. But
changing communication interval also changes the local trajectory, accumulated inner learning
rate, pseudo-gradient scale, and amount of local optimization represented by
one outer update
\citep{douillard2023diloco,kallusky2025snoo}. A fixed outer optimizer then 
become mismatched when communication interval varies. We introduce AutoLoCo, which couples online interval selection with outer optimization adapted to each interval. The controller uses local-drift
energy and aggregation coherence from completed intervals to select a
token-aligned horizon, which is mapped to local steps.
The Outer Optimizer Correction adjusts outer momentum and learning rate by comparing the accumulated inner learning rates of the selected and base intervals. Both intervals are evaluated from the same point in the learning rate schedule. AutoLoCo reduces synchronization rounds by 27\% compared with DiLoCo while achieving a lower mean training loss. The only additional communication consists of scalar summaries and decision metadata.

\textbf{Our contributions} are summarized as follows:
\textbf{(1)} We introduce an online controller that adapts DiLoCo's local interval
    from interval decision. It operates in token-aligned units
    and maps each selected token budget to local steps, which significantly reduces communication frequency and overhead. \textbf{(2)} We introduce an outer optimizer correction for changing local
    horizons. It adjusts outer momentum and learning rate according to the accumulated inner learning rate represented by each interval. \textbf{(3)} We evaluate AutoLoCo across multiple model scales and communication
    settings, and open-source the experimental code and configurations to promote the advancement of LLMs.

\section{Related Work}
\label{sec:related-work}

\textbf{Reducing communication payload.}
Some distributed training works focus on reducing the number of bytes exchanged during each synchronization step through techniques such as quantization, sparsification with error feedback, and structured compression.
QSGD and 1-bit Adam reduce the precision of communicated updates; Deep Gradient Compression and
error feedback transmit selected coordinates while retaining omitted
information; PowerSGD and FetchSGD replace dense updates with low-rank or
sketched representations
\citep{alistarh2017qsgd,tang2021onebitadam,
lin2018deepgradientcompression,karimireddy2019errorfeedback,
vogels2019powersgd,rothchild2020fetchsgd}.
Qsparse-local-SGD combines local computation with quantization and
sparsification, reducing both communication frequency and payload size
\citep{basu2019qsparselocalsgd}.
Recent methods for foundation models adapt these ideas to pretraining and fine tuning over networks with limited bandwidth.
CocktailSGD combines several
compression operators, MuLoCo studies the compressibility of pseudo-gradients under different inner optimizers, and Streaming DiLoCo
communicates parameter subsets while overlapping transfer with local training
\citep{wang2023cocktailsgd,therien2026muloco,douillard2025streaming}.
These methods reduce payload size or exposed communication time.

\textbf{Reducing Communication frequency.}
Local-update methods are performed as multiple optimizer steps between global averages.
FedAvg established this pattern in federated learning, and Local SGD provided
convergence guarantees for periodic averaging
\citep{mcmahan2017fedavg,stich2019localsgd}.
Cooperative SGD and unified Local SGD analyses further connect local work,
communication topology, data heterogeneity, and convergence
\citep{wang2021cooperativesgd,gorbunov2021localsgd}.
Post-local SGD uses a stagewise variant: training begins synchronously and
switches to local updates at a predefined point
\citep{lin2020postlocalsgd}.
DiLoCo brings this structure to language model pretraining through local AdamW,
averaged pseudo-gradients, and a Nesterov outer optimizer; OpenDiLoCo
demonstrates decentralized execution across geographically distributed workers
\citep{douillard2023diloco,jaghouar2024opendiloco}.
HALoS extends low-frequency training to hierarchical asynchronous aggregation
over slow networks, while Scaling Laws for DiLoCo studies examine model size, worker
count, token budget, and optimizer settings
\citep{kim2025halos,charles2025dilocoscaling}.
AutoLoCo, on the other hand, maintains the benefit of infrequent communication, and use dynamic communication interval selected after each interval
from current training behavior instead of a fixed communication clock.

\textbf{Inner-outer optimizer design.}
In local-update training, the inner optimizer determines the trajectory followed
between synchronizations, and the outer optimizer determines how the resulting
pseudo-gradient updates the shared model. SlowMo adds momentum on a slower time
scale after several local steps, and analyses how outer learning
rate, momentum, and acceleration affect Local SGD
\citep{wang2020slowmo,khaled2025outeroptimizers}.
SNOO examines Nesterov momentum applied to pseudo-gradients
\citep{kallusky2025snoo}.
Federated optimization provides related server-side mechanisms. FedOpt applies
adaptive optimizers to aggregated client changes, Mime uses shared optimizer
statistics to preserve centralized optimizer behavior, FedExP adapts the server
step size from update geometry, and FADAS responds to asynchronous staleness
\citep{reddi2021adaptive,karimireddy2021mime,
jhunjhunwala2023fedexp,wang2024fadas}.
FedNova further shows that updates can be normalized according to the effective
local progress performed before aggregation
\citep{wang2020fednova}.
These studies show that the outer optimizer reflect the
optimization process summarized by each aggregated update. AutoLoCo follows
this principle in DiLoCo. We use accumulated inner learning rate to represent
the optimization performed within each interval, then uses this quantity to adjust the outer momentum and learning rate
when the synchronization interval changes.

\textbf{Adaptive synchronization intervals.}
Methods that adapt synchronization intervals during training generally
determine the next interval by following a predefined
schedule or updating the interval from online feedback collected during the
current run. Adaptive synchronization analyses and communication-efficient
SGD derive conditions under which the interval can grow over training
\citep{haddadpour2019adaptive,spiridonoff2021communicationefficient}.
STL-SGD increases the interval when the learning rate enters a new stage, and
the Quadratic Synchronization Rule makes this relationship explicit through
\(H_t \propto 1/\eta_t^2\)
\citep{shen2021stlsgd,gu2024quadratic}.
These schedule-based policies provide simple and predictable interval
trajectories aligned with the optimizer schedule. Feedback-based methods
instead adapt synchronization to observations from the current run. AdaComm
uses an error--runtime model, dynamic model averaging communicates when local
models deviate from a reference, and Adaptive Periodic Averaging adjusts the
next interval from variance across workers
\citep{wang2019adacomm,kamp2018dynamicmodelaveraging,
jiang2020adaptive}.
These studies show that synchronization frequency respond to
optimizer progress. AutoLoCo extends the next interval selection algorithm and how a DiLoCo outer optimizer should interpret
pseudo-gradients produced by intervals that represent different amounts of
local optimization.

AutoLoCo extends interval selection method by treating synchronization timing and outer
optimization as a coupled design problem. 1) We make the synchronization interval responsive to the current training
state. This allows the communication frequency to adapt as the model's
tolerance for longer local intervals changes over the course of training. 2) We interpret intervals of different lengths according to the amount of local optimization they represent. Longer intervals make each pseudo-gradient represent more local
optimization and change its scale, which make each outer step span more
inner optimization and change the effective time unit of the outer momentum and learning rate. Therefore, we combine
adaptive synchronization with outer optimization adapted to each interval, which is a complementary method to compression, quantification, and
asynchronous communication architectures.

\section{Method}
\label{sec:method}

In this section, we introduce our method AutoLoCo, which is refined on the DiLoCo
inner-outer training loop. DiLoCo performs a fixed number of local steps
before averaging pseudo-gradients and applying a Nesterov outer update. We preserves this loop but replaces the
fixed horizon with two adaptive components. The horizon component selects a
token-aligned horizon and maps it to executable local steps, while the outer
correction normalizes the pseudo-gradient and adjusts the outer momentum and
learning rate. These components determine the number of local steps before each
synchronization and align the outer optimizer with the amount of local
optimization represented by the resulting pseudo-gradient.

\begin{figure}[t]
    \centering
    \includegraphics[width=\linewidth,trim=16bp 43bp 25bp 43bp,clip]{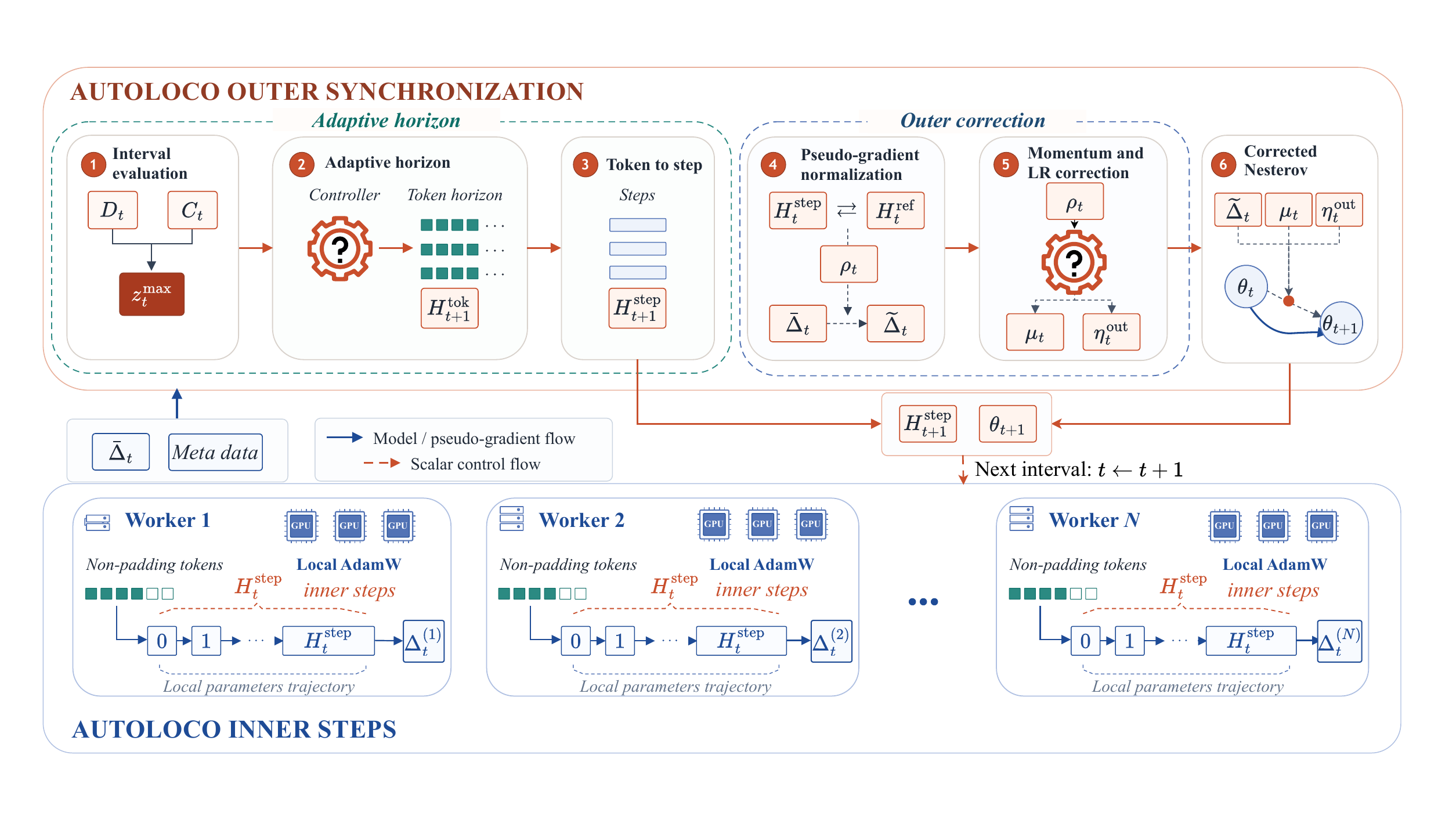}
    \caption{AutoLoCo adapts the local horizon and corrects the outer optimizer in both loops.}
    \label{fig:autoloco-overview}
\end{figure}

\subsection{Adaptive Interval Selection}
\label{sec:adaptive-horizon-selection}

\textbf{Interval Evaluation.}
At the end of interval \(t\), we choose the token-equivalent horizon
\(H_{t+1}^{\mathrm{tok}}\) for the next interval. The controller has the task of  keeping the
accepted horizon, evaluating a larger candidate, or using a smaller horizon. Let
\(N\) be the number of workers, \(\Delta_t^{(i)}\) the pseudo-gradient from
worker \(i\), and
\(\bar{\Delta}_t=N^{-1}\sum_{i=1}^{N}\Delta_t^{(i)}\) their average. We use
\begin{equation}
D_t
=
\frac{1}{N}
\sum\nolimits_{i=1}^{N}
\left\|\Delta_t^{(i)}\right\|_2^2,
\qquad
C_t
=
\frac{
N\left\|\bar{\Delta}_t\right\|_2^2
}{
D_t+\epsilon
},
\label{eq:adaptive-h-core-diagnostics}
\end{equation}
where \(\epsilon>0\) is a numerical constant. \(D_t\) measures the mean
squared displacement from the synchronized model to local
endpoints, while \(C_t\) measures how much remains aligned
after averaging. They separate large but coherent local motion from
movement that cancels across workers. Appendix~\ref{app:dc_geometric_interpretation} derives the relation between
\(D_t\), \(C_t\), and disagreement among local endpoints. We compare both quantities with an
accepted-horizon reference profile estimated from the current run. The profile
stores \(D_t^{\mathrm{ref}}\), \(C_t^{\mathrm{ref}}\), and
\(\Lambda_t^{\mathrm{ref}}\), where
\(\Lambda_t^{\mathrm{ref}}=\sum_{h=1}^{H_t^{\mathrm{step}}}\eta_{t,h}\)
is evaluated on the accepted-horizon interval used to update the profile.
When inner learning rates vary, equal numbers of local steps correspond to different amounts of optimization. STL-SGD and the Quadratic
Synchronization Rule address this by coupling the communication period to the
learning-rate schedule, and FedNova normalizes updates according to
effective local progress
\citep{shen2021stlsgd,gu2024quadratic,wang2020fednova}. Following this
principle, we use accumulated inner learning rate to normalize the drift in
each interval before comparing it with the accepted-horizon profile. We write
\begin{equation}
\xi_t
=
\frac{
\sum_{h=1}^{H_t^{\mathrm{step}}}\eta_{t,h}
}{
\Lambda_t^{\mathrm{ref}}+\epsilon
},
\label{eq:adaptive-h-lr-mass}
\end{equation}
where \(H_t^{\mathrm{step}}\) is the executed step horizon, \(h\) indexes its
local steps, and \(\eta_{t,h}\) is the inner learning rate at step \(h\).
We then write the two departures from the profile directly as
\begin{equation}
\begin{aligned}
r_{D,t}
&=
\log
\left(
\frac{
\sqrt{D_t}
}{
\sqrt{D_t^{\mathrm{ref}}}\,\xi_t+\epsilon
}
+\epsilon
\right),
&
r_{C,t}
&=
-\log
\left(
\frac{
C_t
}{
C_t^{\mathrm{ref}}+\epsilon
}
+\epsilon
\right).
\end{aligned}
\label{eq:adaptive-h-core-residuals}
\end{equation}
Here, \(r_{D,t}\) measures excess drift after accounting for accumulated inner
optimization. \(r_{C,t}\) measures the loss of aggregation coherence relative
to the accepted-horizon profile. Let \(c_{D,t}\) and \(c_{C,t}\) be the
median residuals in the accepted-horizon histories, \(s_{D,t}\) and
\(s_{C,t}\) be their median-absolute-deviation scales.
Appendix~\ref{app:adaptive-h-current-run-reference} explains how these
histories are collected and updated. The controller uses
\begin{equation}
z_t^{\max}
=
\max
\left\{
\frac{[r_{D,t}-c_{D,t}]_+}{s_{D,t}},
\frac{[r_{C,t}-c_{C,t}]_+}{s_{C,t}}
\right\},
\label{eq:adaptive-h-core-deviation}
\end{equation}
where \([x]_+=\max(x,0)\). A small \(z_t^{\max}\) supports a larger
candidate, reference-consistent behavior keeps the accepted horizon,
persistent moderate deviation reduces it, and severe deviation reduces it
immediately. We evaluate a candidate before adoption and re-estimate the
reference profile after adoption or reduction. The controller combines this
assessment with its current phase and horizon bounds to select
\(H_{t+1}^{\mathrm{tok}}\).
Appendix~\ref{app:adaptive-h-candidate-evaluation} provides the assessment thresholds, confirmation procedure, and horizon acceptance or rejection rules.

\textbf{Token to Step Mapping.}
We control local work in non-padding tokens rather than raw steps because
padding, packing, and minibatch composition can change the training volume of
one optimizer step. Here, \(n_{t,h}\) is the global non-padding token count at
local step \(h\), and \(M_t\) is the sum of these counts over interval \(t\).
\(H_{\mathrm{base}}\) is the recipe token horizon, and
\(M_{\mathrm{base}}\) is the median token mass of current-run bootstrap
intervals executed at \(H_{\mathrm{base}}\). After each interval, we compute
\(\bar n_t=M_t/H_t^{\mathrm{step}}\) and update
\(\hat n_t=\beta\hat n_{t-1}+(1-\beta)\bar n_t\), where \(\hat n_t\) is the
exponential moving average of tokens per executable step and \(\beta\) is its
coefficient. Before rounding, the mapped step horizon is
\begin{equation}
H_t^{\mathrm{step}}
\approx
\frac{
M_{\mathrm{base}}H_t^{\mathrm{tok}}
}{
H_{\mathrm{base}}(\hat n_{t-1}+\epsilon)
}.
\label{eq:adaptive-h-token-mapping}
\end{equation}
We round this value to the nearest admissible multiple of the horizon granularity \(q_H\), using the procedure in
Appendix~\ref{app:adaptive-h-token-control}. Applying the same mapping with
\(H_t^{\mathrm{tok}}=H_{\mathrm{base}}\) gives \(H_t^{\mathrm{ref}}\), the
executable horizon for one recipe token interval at the current token density.

\subsection{Outer Optimizer Correction for Variable Local Horizons}
\label{sec:outer-optimizer-correction}

Variable horizons change the meaning of an outer update. A longer interval
alter the scale of the averaged pseudo-gradient and make one momentum update
span more inner optimization. With a fixed outer optimizer, it applies the same
update rule to every synchronization. Prior work on slow outer momentum,
DiLoCo Nesterov updates, normalization under unequal local progress, and
adaptive server optimization shares the principle: the outer update
should align with the local computation that produced it
\citep{wang2020slowmo,kallusky2025snoo,wang2020fednova,
reddi2021adaptive}. We apply this principle across successive DiLoCo
intervals. Our outer correction compares the selected interval with one base
token interval at the same position in the inner learning-rate schedule, then
uses the resulting ratio to normalize the pseudo-gradient, adjust momentum
retention, and bound the outer-step scale. A detailed description of this mismatch
appears in
Appendix~\ref{app:outer-correction-motivation}.

\textbf{Interval ratio and pseudo-gradient normalization.}
For interval \(t\), \(H_t^{\mathrm{step}}\) is the selected step
horizon, and \(H_t^{\mathrm{ref}}\) is the horizon corresponding to
one base token interval at the current token density. Let \(h\) index local
steps, let \(\eta_{t,h}\) be the inner learning rate at step \(h\), and let
\(\epsilon>0\) be a numerical constant. We preview both horizons from the same
scheduler position and define
\begin{equation}
\rho_t
=
\max
\left(
\frac{
\sum_{h=1}^{H_t^{\mathrm{step}}}\eta_{t,h}
}{
\sum_{h=1}^{H_t^{\mathrm{ref}}}\eta_{t,h}
+\epsilon
},
1
\right).
\label{eq:outer-correction-ratio}
\end{equation}
The interval ratio \(\rho_t\) measures the accumulated inner learning rate of
the selected interval in base-interval units. After workers complete the
interval and average their pseudo-gradients, we set
\begin{equation}
\widetilde{\Delta}_t
=
\frac{\bar{\Delta}_t}{\rho_t},
\label{eq:normalized-pseudo-gradient}
\end{equation}
where \(\bar{\Delta}_t\) is the averaged pseudo-gradient and
\(\widetilde{\Delta}_t\) is the normalized outer-optimizer input. At
\(\rho_t=1\), the pseudo-gradient remains unchanged. For
\(\rho_t>1\), the normalization expresses it on the scale of one base
interval. Appendix~\ref{app:outer-correction-interval-comparison} details the
scheduler-aligned comparison, and
Appendix~\ref{app:outer-correction-normalization} describes how the
normalization affects the DiLoCo update. Appendix~\ref{app:lr_mass_interval_ratio} provides a local displacement
interpretation of this ratio and bounds the approximation error. For experiment settings, C4 pretraining does not use normalization, while 8B fine-tuning uses normalization.  Section~\ref{sec:component-ablation} discusses
this difference and presents the corresponding ablation results.

\textbf{Momentum and outer-step adjustment.}
Normalization corrects the optimizer input, but a longer interval also changes
how long the incoming momentum state persists in units of inner optimization.
Let \(\mu_{\mathrm{base}}\) denote the base outer momentum. We first set
\(\mu_t=\mu_{\mathrm{base}}^{\rho_t}\), then constrain the
interval-adjusted momentum \(\mu_t\) to the configured range
\([\mu_{\min},\mu_{\max}]\). We next set
\(\kappa_t=\min(\rho_t,\kappa_{\max})\), where \(\kappa_t\) is the bounded
outer-step scale and \(\kappa_{\max}\) is its upper limit. Given the base
outer learning rate \(\eta_{\mathrm{base}}^{\mathrm{out}}\), we use
\begin{equation}
\eta_t^{\mathrm{out}}
=
\eta_{\mathrm{base}}^{\mathrm{out}}
\,
\kappa_t
\,
\frac{1-\mu_t}{1-\mu_{\mathrm{base}}},
\label{eq:adjusted-outer-learning-rate}
\end{equation}
where \(\eta_t^{\mathrm{out}}\) is the outer learning rate applied to
interval \(t\). The same \(\rho_t\) controls pseudo-gradient normalization,
momentum retention, and bounded scaling of the outer step. The
derivation is given in Appendix~\ref{app:outer_correction_derivation},
and the reference behavior is described in
Appendix~\ref{app:outer-correction-momentum-step}. The experiment
settings are reported in Table~\ref{tab:exp_controller_settings}.

\subsection{AutoLoCo: Adaptive Gradient Synchronization Meets Outer
Optimizer Correction}
\label{sec:autoloco-coupled-update}

AutoLoCo couples synchronization timing with outer optimization
through a feedback loop that operates at the interval level. At the start of interval
\(t\), the token mapper converts the token-equivalent horizon
\(H_t^{\mathrm{tok}}\), selected from the previous completed interval, into
\(H_t^{\mathrm{step}}\), the executable local-step horizon, and
\(H_t^{\mathrm{ref}}\), the executable horizon corresponding to one base
token interval at the current token density. These horizons determine
\(\rho_t\) and the interval-adjusted outer optimizer settings before local
execution. Each worker starts from \(\theta_t\), follows a local trajectory of
length \(H_t^{\mathrm{step}}\), and contributes a pseudo-gradient. After
synchronization, we normalize the averaged pseudo-gradient and apply the
corrected outer update. Once this update produces \(\theta_{t+1}\), we compute
\(D_t\), \(C_t\), and the completed token statistics. These quantities update
the accepted-horizon profile and token mapper, and determine
\(H_{t+1}^{\mathrm{tok}}\). The horizon component then sets the local
trajectory. The outer correction determines how the resulting
pseudo-gradient updates the synchronized model.

The two components meet in the corrected Nesterov outer update used by
DiLoCo. Let \(v_{t-1}\) be the outer momentum
state entering interval \(t\), \(v_t\) its updated value, and \(d_t\) the
Nesterov update direction. We compute
\begin{equation}
v_t=\mu_t v_{t-1}+\widetilde{\Delta}_t,
\qquad
d_t=\widetilde{\Delta}_t+\mu_t v_t,
\qquad
\theta_{t+1}=\theta_t-\eta_t^{\mathrm{out}}d_t.
\label{eq:autoloco-corrected-nesterov}
\end{equation}
Here, \(\theta_t\) and \(\theta_{t+1}\) are the synchronized parameters
before and after the outer update. The normalized pseudo-gradient
\(\widetilde{\Delta}_t\) provides the optimizer input,
\(\mu_t\) controls momentum retention, and
\(\eta_t^{\mathrm{out}}\) scales the parameter update.

Algorithm~\ref{alg:autoloco-outer-interval} in
Appendix~\ref{app:algorithm} traces one complete AutoLoCo interval.
Appendices~\ref{sec:diloco-preliminaries} and~\ref{sec:fixed-horizon-special-case}
give the inherited DiLoCo update and the conditions for recovering DiLoCo,
respectively.

\begin{table}[!b]
\centering
\caption{Quality and communication cost at step 50,000.
Training loss averages eight workers over the final 1,000 steps;
validation evaluates the final synchronized checkpoint.
Payload is measured in full-model BF16 tensors per worker.
$\dagger$ counts gradient synchronizations; other counts are outer
synchronizations.
$\ddagger$ Given the differences in training settings, we retain their
proxies for selecting communication intervals and adapt the methods
to LLM training.
Bold and underline indicate the best and second-best results,
respectively, in the loss and synchronization count columns.}
\label{tab:pretrain-quality-communication}
\label{tab:exp_main_results}
\vspace{\baselineskip}
\begin{tabular*}{\linewidth}{@{\extracolsep{\fill}}lrrrr@{}}
\toprule
Method
& \shortstack{Training\\loss $\downarrow$}
& \shortstack{Number of\\syncs $\downarrow$}
& \shortstack{Payload\\per sync}
& \shortstack{Validation\\loss $\downarrow$} \\
\midrule
DDP
& \textbf{2.7227} & $50{,}000^{\dagger}$ & $1\times$
& \textbf{2.7233} \\
DiLoCo
& 2.8004 & 100 & $1\times$ & 2.8020 \\
Linear Interval$^{\ddagger}$
& 2.8706 & 100 & $1\times$ & 2.8694 \\
QSR$^{\ddagger}$
& 2.8914 & $\textbf{41}$ & $1\times$ & 2.8793 \\
\midrule
\textbf{AutoLoCo}
& \underline{2.7843} & \underline{73} & $1\times$
& \underline{2.7842} \\
\bottomrule
\end{tabular*}
\end{table}

\section{Experiments}
\label{sec:experiments}

We evaluate AutoLoCo with LLaMA style configuration by 215M total parameters ~\citep{touvron2023llama} and Llama-3.1-8B ~\citep{dubey2024llama3}.
We pre-train on C4~\citep{raffel2020exploring} for 50,000 optimizer steps and match this budget across methods, following the controlled comparisons of \citet{jaghouar2024opendiloco}.
Our main setup uses 8 workers with batch size of 128 per worker, giving a global batch of 1,024.
With a sequence length of 1,024, each step processes 1.05M input token positions.
To test AutoLoCo on a larger model, we full-parameter fine-tune Llama-3.1-8B for 1 epoch on UltraChat-200k~\citep{ding2023enhancing} with 8 and 16 workers.
For the experiment of 16 workers, we set up actual distributed workers across different hosts.
Both settings use a global batch of 32 and a maximum sequence length of 2,048.
In the 16 workers setup, the effective throughput measured per-worker ranges from 1.26 to 1.46~Gbit/s.
Appendix~\ref{app:experimental_settings} reports the implementation settings, data processing and evaluation protocol.
Appendix~\ref{app:int8-communication} provides implementation details and additional experimental results on combining AutoLoCo with low-precision communication.

\subsection{Pretrain Quality and Communication Cost}
\label{sec:exp_main_results}
\label{sec:pretrain-quality-communication}

We compared with two strong baselines, data-parallel training (DDP)~\citep{li2020pytorchdistributed} and DiLoCo. In the main setup, DiLoCo follows the configuration
reported in prior work. AutoLoCo retains
most settings and adjusts the parameters of its adaptive
components. AutoLoCo uses 73 synchronizations compared with 100 for
DiLoCo, while training loss improves from
2.8004 to 2.7843 and validation loss from 2.8020 to 2.7842
(Table~\ref{tab:pretrain-quality-communication}). With the same payload
per synchronization, this saves 27\% of the cumulative logical volume. Appendix~\ref{app:experimental_data_budget} gives the protocol, and
Appendix~\ref{app:pretrain-outer} examines the effect of the outer
settings under identical intervals.

\begin{figure}[!htbp]
\centering
\begin{minipage}[t]{0.49\linewidth}
\centering
\includegraphics[width=\linewidth,trim=0bp 5bp 0bp 0bp,clip]{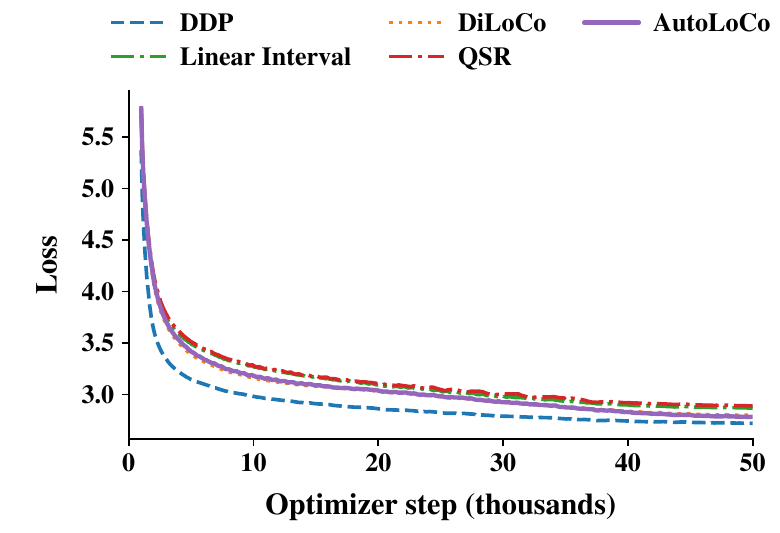}
\par (a) Full trajectory
\end{minipage}\hfill
\begin{minipage}[t]{0.49\linewidth}
\centering
\includegraphics[width=\linewidth,trim=0bp 5bp 0bp 0bp,clip]{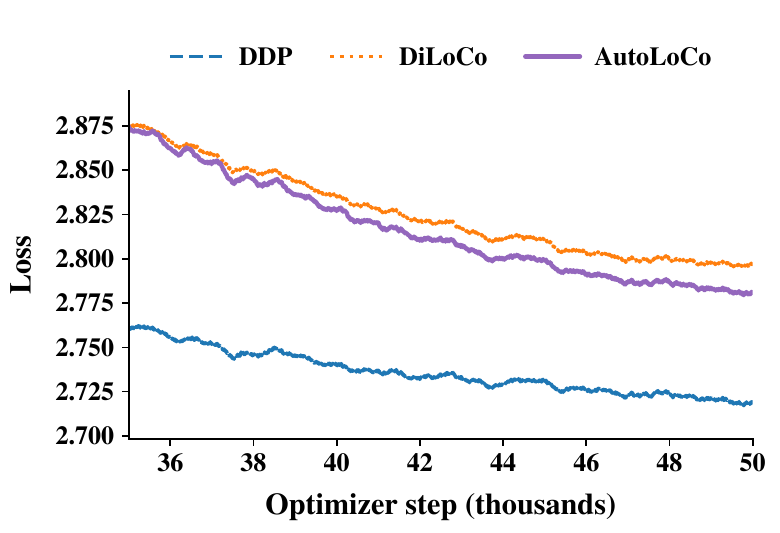}
\par (b) Final portion
\end{minipage}
\caption{Training loss over the full budget and during the
final portion of training.}
\label{fig:pretrain-optimization}
\label{fig:pretrain-tail}
\end{figure}

For pre-training, we also compared the Quadratic Synchronization Rule (QSR)~\citep{gu2024quadratic} and the linearly increasing communication-interval schedule of \citet{spiridonoff2021communicationefficient}, which we call Linear Interval. Our local
AdamW implementations of Linear
Interval and
QSR follow the prior works, use direct model averaging. AutoLoCo finishes with better
validation loss than both. DDP remains the quality reference, reaching validation loss
2.7233 with gradient synchronization at every step. The training loss advantage over DiLoCo emerges late
(Figure~\ref{fig:pretrain-optimization}). 

The threshold comparison reflects this change: AutoLoCo reaches 2.90
slightly later, but 2.85 and 2.80 earlier
(Table~\ref{tab:pretrain-threshold-times} and Figure~\ref{fig:pretrain-thresholds}). The results in the table also show that AutoLoCo achieves a more competitive rate of loss reduction in the later stages of training than DiLoCo and the other two methods. The analysis in Appendix~\ref{app:pretrain-trajectories} shows the same late
advantage. The communication savings are clearer than the change in total runtime.
Observed interface I/O falls by 26.5\%, and
accumulated synchronization boundary time by 12.9\%, from 40.74 to
35.50 minutes. Our diagnostic and control accounting totals 0.304\% of
training time (Table~\ref{tab:pretrain-control-overhead}). This includes preparation work and auxiliary diagnostics alongside the
controller. Control endpoint accounting totals 2.89~MiB for the run.

\begin{table}[htbp]
\centering
\caption{Cumulative recorded step time (hours) to first reach each
threshold of the 8 worker mean loss, using a 1,000-step trailing average.
A dash indicates that the threshold is not reached within 50,000 steps.
The best time in each column is shown in bold.}
\label{tab:pretrain-threshold-times}
\vspace{\baselineskip}
\begin{tabular*}{\linewidth}{@{\extracolsep{\fill}}lrrrrrr@{}}
\toprule
Loss threshold & 3.50 & 3.20 & 3.00 & 2.90 & 2.85 & 2.80 \\
\cmidrule(l){2-7}
Method & \multicolumn{6}{c}{Cumulative recorded step time (h) $\downarrow$} \\
\midrule
DiLoCo
& 1.31
& 2.83
& 7.48
& 10.60
& 12.44
& 15.26 \\
Linear Interval
& 1.68
& 4.34
& 9.13
& 12.87
& \multicolumn{1}{c}{--}
& \multicolumn{1}{c}{--} \\
QSR
& 1.69
& 4.30
& 9.40
& 14.27
& \multicolumn{1}{c}{--}
& \multicolumn{1}{c}{--} \\
AutoLoCo
& 1.39
& 3.03
& 7.52
& 10.61
& \textbf{12.09}
& \textbf{14.22} \\
\bottomrule
\end{tabular*}
\end{table}

\subsection{Fine-Tuning Quality and Communication Cost}
\label{sec:ultrachat-sft}

We fine-tune Llama-3.1-8B on UltraChat for 1 epoch with
a global batch of 32. Both 8 workers and 16 workers experiments share this setting. DiLoCo follows the experimental configuration
reported in prior work. Our AutoLoCo
recipe uses similar hyper parameter compared with main setup. The only difference is that we apply pseudo-gradient normalization
in the 8B experiments. Section~\ref{sec:component-ablation} discusses
this difference and presents the corresponding ablation results. Appendices~\ref{app:experimental_data_budget} and~\ref{app:ultrachat-protocol}
give the training and evaluation protocol.

\begin{table}[!ht]
\centering


\begin{minipage}[t]{0.52\linewidth}
\caption{UltraChat training and communication with eight workers, a global
batch of 32 for 1 epoch. Training loss covers the final
1,000 steps. Logical payload represents the volume of model data exchanged
at the algorithmic level. Actual traffic over physical links depends
on the distributed framework and the collective communication algorithm.}
\label{tab:ultrachat_8gpu_training}
\end{minipage}\hfill
\begin{minipage}[t]{0.46\linewidth}
\caption{UltraChat training and communication with 16 workers, a global
batch of 32 for 1 epoch. Both methods communicate BF16
pseudo-gradients. Logical payload represents the volume of model data exchanged
at the algorithmic level. Actual traffic over physical links depends
on the distributed framework and the collective communication algorithm. Metric definitions follow
Table~\ref{tab:ultrachat_8gpu_training}.}
\label{tab:ultrachat_16worker_training}
\end{minipage}

\par
\vspace{0.5\baselineskip}


\noindent
\begin{minipage}[t]{0.52\linewidth}
\vspace{0pt}
\centering
\setlength{\tabcolsep}{2pt}
\begin{tabular*}{\linewidth}{@{\extracolsep{\fill}}lrrr@{}}
\toprule
Method & DDP & DiLoCo & AutoLoCo \\
\midrule
Train loss $\downarrow$
& \tikzmarknode{t3first}{\strut 0.8736}
& 0.9102
& \textbf{0.8759} \\
Syncs $\downarrow$
& 6,490
& 65
& 49 \\
\multicolumn{4}{c}{} \\
\midrule
\multicolumn{4}{@{}l}{Logical payload (GB)} \\
Per sync
& \multicolumn{1}{c}{--}
& 16.061
& 16.061 \\
Total $\downarrow$
& \multicolumn{1}{c}{--}
& 1,043.934
& \textbf{786.966} \\
\midrule
\multicolumn{4}{@{}l}{UltraChat test} \\
NLL $\downarrow$
& 0.8745
& 0.9112
& \textbf{0.8766} \\
PPL $\downarrow$
& \tikzmarknode{t3last}{\strut 2.3976}
& 2.4874
& \textbf{2.4027} \\
\bottomrule
\end{tabular*}

\begin{tikzpicture}[remember picture,overlay]
\draw[
    line width=0.7pt,
    dash pattern=on 2.5pt off 1.3pt
]
([xshift=\tabcolsep,yshift=\belowrulesep]t3first.north east)
--
([xshift=\tabcolsep,yshift=-\aboverulesep]t3last.south east);
\end{tikzpicture}
\end{minipage}\hfill
\begin{minipage}[t]{0.46\linewidth}
\vspace{0pt}
\centering
\setlength{\tabcolsep}{2pt}
\begin{tabular*}{\linewidth}{@{\extracolsep{\fill}}lrr@{}}
\toprule
Method & DiLoCo & AutoLoCo \\
\midrule
Train loss $\downarrow$ & 0.9003 & \textbf{0.8761} \\
Syncs $\downarrow$ & 65 & \textbf{49} \\
Wall time (h) $\downarrow$ & 7.2312 & \textbf{6.4958} \\
\midrule
\multicolumn{3}{@{}l}{Logical payload (GB)} \\
Per sync & 16.061 & 16.061 \\
Total $\downarrow$ & 1,043.934 & \textbf{786.966} \\
\midrule
\multicolumn{3}{@{}l}{UltraChat test} \\
NLL $\downarrow$ & 0.9016 & \textbf{0.8772} \\
PPL $\downarrow$ & 2.4635 & \textbf{2.4041} \\
\bottomrule
\end{tabular*}
\end{minipage}

\par\medskip


\noindent
\begin{minipage}[t]{0.52\linewidth}
\caption{Downstream evaluation after UltraChat SFT with eight workers.
Underline marks the next highest
distinct score. Evaluation settings are given in
Appendix~\ref{app:ultrachat-protocol}.}
\label{tab:ultrachat_sft_8workers}
\end{minipage}\hfill
\begin{minipage}[t]{0.46\linewidth}
\caption{Downstream evaluation after UltraChat SFT with 16 workers.
Scores are percentages; higher is better. Bold marks the best observed
score in each row. Evaluation settings match
Table~\ref{tab:ultrachat_sft_8workers}.}
\label{tab:ultrachat_sft_16workers}
\end{minipage}

\par
\vspace{0.5\baselineskip}


\noindent
\begin{minipage}[t]{0.52\linewidth}
\vspace{0pt}
\centering
\setlength{\tabcolsep}{2pt}
\begin{tabular*}{\linewidth}{@{\extracolsep{\fill}}lrrr@{}}
\toprule
Metric & DDP & DiLoCo & AutoLoCo \\
\midrule
IFEval Strict $\uparrow$
& \tikzmarknode{t5first}{\strut 37.15}
& 32.72
& \textbf{32.90} \\
IFBench Loose $\uparrow$
& 17.33
& 16.00
& \textbf{17.33} \\
MMLU-Pro Accuracy $\uparrow$
& 36.54
& 31.21
& \textbf{36.25} \\
BBH CoT EM $\uparrow$
& 61.79
& 56.03
& \textbf{61.10} \\
HumanEval+ $\uparrow$
& \tikzmarknode{t5last}{\strut 25.61}
& 15.24
& \textbf{22.56} \\
\bottomrule
\end{tabular*}

\begin{tikzpicture}[remember picture,overlay]
\draw[
    line width=0.7pt,
    dash pattern=on 2.5pt off 1.3pt
]
([xshift=\tabcolsep,yshift=\belowrulesep]t5first.north east)
--
([xshift=\tabcolsep,yshift=-\aboverulesep]t5last.south east);
\end{tikzpicture}
\end{minipage}\hfill
\begin{minipage}[t]{0.46\linewidth}
\vspace{0pt}
\centering
\setlength{\tabcolsep}{2pt}
\begin{tabular*}{\linewidth}{@{\extracolsep{\fill}}lrr@{}}
\toprule
Metric & DiLoCo & AutoLoCo \\
\midrule
IFEval Strict $\uparrow$ & 34.01 & \textbf{34.38} \\
IFBench Loose $\uparrow$ & 17.33 & \textbf{17.67} \\
MMLU-Pro Accuracy $\uparrow$ & 33.10 & \textbf{36.23} \\
BBH CoT EM $\uparrow$ & 57.93 & \textbf{62.71} \\
HumanEval+ $\uparrow$ & 17.07 & \textbf{23.17} \\
\bottomrule
\end{tabular*}
\end{minipage}

\end{table}

With eight workers, AutoLoCo reduces synchronizations from 65 to 49 and
logical payload by 24.6\% relative to DiLoCo. Test NLL improves from
0.9112 to 0.8766, leaving a gap of 0.0021 to DDP
(Table~\ref{tab:ultrachat_8gpu_training}). For 16 workers. AutoLoCo reaches
test NLL 0.8772 compared with DiLoCo's 0.9016, while retaining 49
synchronizations and the 24.6\% payload reduction
(Table~\ref{tab:ultrachat_16worker_training}). 

With eight workers, AutoLoCo improves on DiLoCo by about five
percentage points on MMLU-Pro and BBH, and by 7.32 points on
HumanEval+ (Table~\ref{tab:ultrachat_sft_8workers}). The gap to DDP is
below 0.7 points on MMLU-Pro and BBH. IFEval improves by just 0.18
points over DiLoCo and remains 4.25 points behind DDP. With 16 workers, AutoLoCo scores above DiLoCo on all five tasks
(Table~\ref{tab:ultrachat_sft_16workers}). The largest gain is again
in code generation: HumanEval+ improves by 6.10 percentage points,
followed by BBH at 4.78 points. Both instruction-following scores
improve by less than half a point.

\subsection{Ablation}
\label{sec:component-ablation}

We examine how adaptive synchronization and outer correction contribute
to AutoLoCo on C4 with 4 workers. All runs share the initialization,
training data, inner AdamW recipe, and 50,000 step budget, with a global
batch of 512. Appendix~\ref{app:component-ablation} gives the component
settings.

\begin{table}[!ht]
\centering
\caption{Component ablation on C4 with 4 workers and 50,000 optimizer
steps. Adaptive intervals include
horizon selection and token mapping. Bold marks the best training loss.}
\label{tab:component-ablation}
\setlength{\tabcolsep}{4pt}
\vspace{\baselineskip}
\begin{tabular*}{\linewidth}{@{\extracolsep{\fill}}lcccrrr@{}}
\toprule
& & \multicolumn{2}{c}{Outer correction} & & & \\
\cmidrule(lr){3-4}
Variant
& \shortstack{Adaptive\\intervals}
& Momentum
& \shortstack{Learning rate}
& \shortstack{Training\\loss $\downarrow$}
& Syncs $\downarrow$
& \shortstack{Training PPL} \\
\midrule
DiLoCo
& -- & -- & --
& 2.8251 & 100 & 16.8618 \\
Adaptive intervals
& $\checkmark$ & -- & --
& 2.8310 & 73 & 16.9618 \\
Adaptive + momentum
& $\checkmark$ & $\checkmark$ & --
& 2.8484 & 73 & 17.2603 \\
AutoLoCo
& $\checkmark$ & $\checkmark$ & $\checkmark$
& \textbf{2.8116} & \textbf{73} & \textbf{16.6362} \\
\bottomrule
\end{tabular*}
\end{table}

Adaptive intervals alone save communication but finish with a training
loss of 2.8310, slightly above DiLoCo's 2.8251. Adjusting momentum alone
increases the loss to 2.8484. During the $H=750$ phase, the median outer
update magnitude is 36.45\% smaller than with adaptive intervals alone,
consistent with a more conservative outer response. Adding the outer-step adjustment reverses this loss increase. AutoLoCo
reaches 2.8116, an improvement of 0.0368 over the variant with momentum
adjustment alone and 0.0135 over DiLoCo. These results support our design
of adjusting momentum retention and outer-step scaling together as
local intervals change.

\begin{table}[!ht]
\centering
\begin{minipage}[t]{0.48\linewidth}
\centering
\caption{Normalization in C4 pretraining (4 workers)}
\label{tab:ablation-normalization}
\vspace{\baselineskip}
\begin{tabular*}{\linewidth}{@{\extracolsep{\fill}}lrr@{}}
\toprule
& \multicolumn{2}{c}{Normalization} \\
\cmidrule(lr){2-3}
Metric & Without & With \\
\midrule
Training loss & 2.811582 & 2.826429 \\
Syncs & 73 & 73 \\
Training time (h) & 23.1001 & 23.1300 \\
\bottomrule
\end{tabular*}
\end{minipage}\hfill
\begin{minipage}[t]{0.48\linewidth}
\centering
\caption{Normalization in fine-tuning(16 workers).}
\label{tab:ultrachat_normalization_ablation}
\vspace{\baselineskip}
\begin{tabular*}{\linewidth}{@{\extracolsep{\fill}}lrr@{}}
\toprule
& \multicolumn{2}{c}{Normalization} \\
\cmidrule(lr){2-3}
Metric & Without & With \\
\midrule
Training loss & 0.876554 & 0.876079 \\
Syncs & 49 & 49 \\
Test NLL & 0.877551 & 0.877184 \\
Test PPL & 2.405003 & 2.404119 \\
\bottomrule
\end{tabular*}
\end{minipage}

\end{table}

\paragraph{Pseudo-gradient normalization.}
Normalization has different effects across the two training settings. For fixed coefficients and a constant input, normalization changes
the Nesterov steady-state response coefficient to the unnormalized
averaged pseudo-gradient from
\(\eta_t^{\mathrm{out}}/(1-\mu_t)\) to
\(\eta_t^{\mathrm{out}}/[\rho_t(1-\mu_t)]\).
Under our outer correction, this cancels the additional amplification
by \(\rho_t\) when the upper bound on outer-step scaling is inactive. With identical horizon trajectories and outer coefficients,
normalization reduces the median outer-update magnitude by 47.50\%
during the \(H=750\) phase in C4, but increases final training loss.
In the reported SFT comparison, normalization slightly lowers test NLL. This observation is consistent with a more conservative outer update, but does not by itself establish better retention of pretrained capabilities. These results motivate different normalization settings for the two evaluated training recipes.
Therefore, whether to apply pseudo-gradient normalization should depend on the task.

\section{Conclusion}
\label{sec:conclusion}

We introduced AutoLoCo, which adapts synchronization intervals and
corrects the outer optimizer for variable local horizons.
Our pretraining and fine-tuning experiments show that AutoLoCo reduces
communication frequency while preserving training performance.
Relative to DiLoCo, the reductions are 27\% in pretraining and
24.6\% in fine-tuning.
Beyond the precision settings evaluated here, future work could
explore lower precision communication with error feedback to
compensate for quantization error.
Another direction is to extend the adaptive range to longer
communication intervals and study their effects on training
performance and stability.

\bibliography{iclr2027_conference}
\bibliographystyle{iclr2027_conference}

\appendix
\section{Implementation details for adaptive horizon selection}
\label{app:adaptive-horizon-details}

\subsection{Current Run Reference}
\label{app:adaptive-h-current-run-reference}

We estimate the accepted-horizon reference profile from the current run. We keep the recipe horizon throughout the 1,000 step
inner-learning-rate warm-up, which spans the first two 500 step intervals.
Beginning with the interval that completes warm-up, we collect two finite
intervals at the initial horizon and set the reference statistics to the
geometric means of their observations. We then keep the same horizon for three
additional intervals to estimate the normal variation of \(r_{D,t}\) and
\(r_{C,t}\). For each residual family, the robust center is
the median of at most 16 stored residuals. We first assess each completed interval and then add its residuals to the
history only when the interval is increase supporting or reference
consistent.
After calibration, only increase-supporting or reference-consistent intervals
executed at the accepted horizon update the residual history. The two
diagnostic references use a log-space exponential moving average, while \(\Lambda_t^{\mathrm{ref}}\) follows the latest accepted
interval because the inner learning-rate schedule changes over training.

\subsection{Interval Evaluation}
\label{app:adaptive-h-candidate-evaluation}
\textbf{Monitoring the accepted horizon.}
As in Table~\ref{tab:adaptive-h-assessment}, a reference-consistent interval keeps the accepted horizon. A larger horizon is considered only after the post-reduction hold-off ends and at least one interval at the accepted horizon is consistent with the reference profile.
A moderate deviation keeps the current horizon, but two consecutive moderate deviations trigger a horizon reduction.
A severe deviation reduces
the horizon immediately. A non-finite required statistic uses the same
reduction path and may invoke the minimum-horizon fallback.

\begin{table}[!ht]
\centering
\caption{Interval assessments for the active \(D_t\) and \(C_t\) classifier in
the pre-training experiments.}
\label{tab:adaptive-h-assessment}
\vspace{\baselineskip}
\begin{tabular}{@{}ll@{}}
\toprule
\textbf{Interval assessment} & \textbf{Range of \(z_t^{\max}\)} \\
\midrule
Increase supported
& \(z_t^{\max}<1.5\) \\
Reference consistent
& \(1.5\le z_t^{\max}<2.5\) \\
Moderate deviation
& \(2.5\le z_t^{\max}<4.0\) \\
Severe deviation
& \(z_t^{\max}\ge4.0\) \\
\bottomrule
\end{tabular}
\end{table}

\textbf{Selecting a longer interval.}
When an increase is supported, we search the admissible
horizons above the accepted value. Each larger horizon defines a longer
interval and is mapped to steps before evaluation. We preview the
accumulated inner learning rate that the interval would cover at the current
scheduler position and compare that exposure with the accepted horizon. We choose the longest admissible interval whose previewed
exposure is no more than \(1.15\) times that of the accepted horizon. 

A previously rejected interval leaves its token-equivalent horizon as a
rejected-horizon bound. While that bound is active, the next interval uses the
nearest admissible horizon to the geometric midpoint between the accepted
horizon and the rejected bound. The controller also stores the accumulated
inner learning rate observed when the bound was created. If a later preview at
the rejected bound falls to at most \(0.90\) of that stored value, we retire
the bound because learning-rate decay has changed the amount of local
optimization represented by the same horizon.

\begin{figure}[t]
    \centering
    \includegraphics[width=\linewidth]{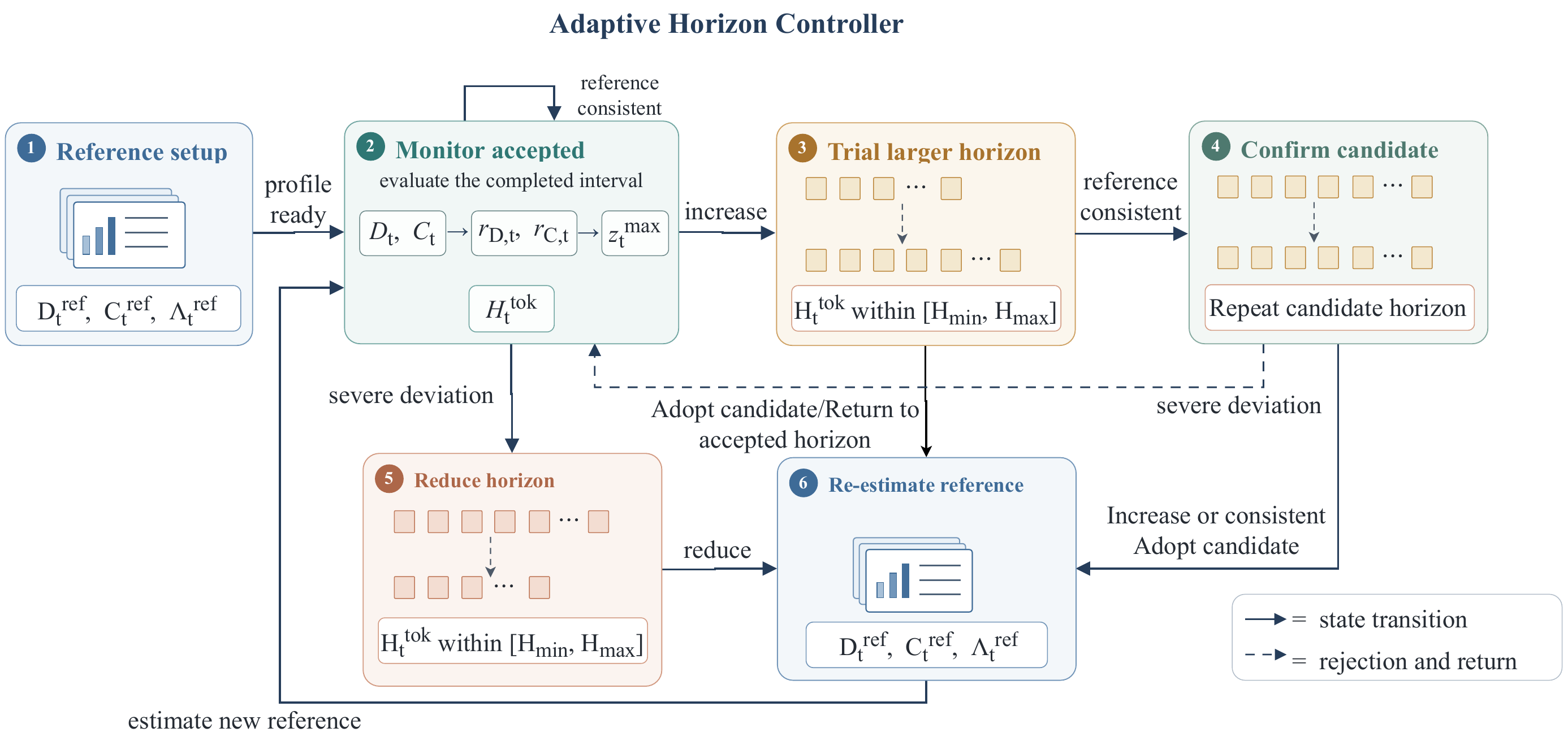}
    \caption{Adaptive horizon selection in AutoLoCo.
A longer candidate is tested only after the reference and monitoring
requirements are met. A first candidate interval that supports an increase
leads to immediate adoption, whereas a reference-consistent result
requires one additional interval. Dashed intermodule arrows denote
candidate rejection and return to the previously accepted horizon.
The reference profile is re-estimated after adoption, return, or reduction.
Numerical safeguards are omitted.}
    \label{fig:adaptive-h-controller-flow}
\end{figure}

\textbf{Evaluating the longer interval.}
We evaluate the first interval executed at a larger candidate horizon.
If it supports an increase, we adopt the candidate immediately. If it is
reference consistent, we retain the same candidate token horizon for one
additional interval and adopt it if that interval supports an increase
or is reference consistent. A moderate or severe deviation in either
interval rejects the candidate and restores the previously accepted
horizon. After adoption or return, we re-estimate the reference profile
at the horizon used next before resuming monitoring.
Figure~\ref{fig:adaptive-h-controller-flow} summarizes these transitions.

\subsection{Horizon Updates}
\label{app:adaptive-h-token-control}

\textbf{Token mapping calibration.}
The token mapper is calibrated independently of the diagnostic profile. During
the first three recipe-horizon intervals, we record the exact global
non-padding token count and set \(M_{\mathrm{base}}\) to their median. We
initialize the tokens-per-step estimate from the first completed interval and
then update \(\hat n_t\) after every interval with \(\beta=0.9\). The newly
selected \(H_{t+1}^{\mathrm{tok}}\) is mapped only after this update, so the next
interval uses the latest observed token density. No historical trace or
external fixed-horizon run supplies \(M_{\mathrm{base}}\).

Before \(M_{\mathrm{base}}\) and \(\hat n_t\) are available, we temporarily
execute the same numerical value for \(H_t^{\mathrm{tok}}\) and
\(H_t^{\mathrm{step}}\). After calibration, we evaluate the ratio in
Equation~(\ref{eq:adaptive-h-token-mapping}) and choose the nearest admissible
multiple of \(q_H\). The interval \([H_{\min},H_{\max}]\) constrains the
token-equivalent controller domain. Applying the same mapping at
\(H_t^{\mathrm{tok}}=H_{\mathrm{base}}\) gives \(H_t^{\mathrm{ref}}\), which
the outer correction uses as the current-density version of one recipe token
interval.

\textbf{Reducing the horizon.}
Under normal training dynamics, \(H\) remains stable or gradually increases;
the reduction rule is used to handle abnormal or persistently adverse
observations. A confirmed moderate deviation reduces the current
token-equivalent horizon by a factor of \(1.20\), while a severe deviation
reduces it by a factor of \(1.50\). We then select the nearest admissible
multiple of \(q_H\) that is strictly smaller than the current horizon and lies
within \([H_{\min},H_{\max}]\). When a larger candidate is rejected, the
controller normally returns to the previously accepted horizon rather than
applying an additional multiplicative reduction. Repeated severe or invalid
observations may trigger the \(H_{\min}\) fallback. After a reduction, one
interval at the new horizon initializes its reference profile, three
additional intervals estimate reference variability, and one monitoring
interval must complete before the controller can propose another increase.

\section{Implementation Details for the Outer Optimizer Correction}
\label{app:outer-optimizer-correction-details}

\subsection{Why Variable Horizons Change the Outer Update}
\label{app:outer-correction-motivation}

DiLoCo processes its outer optimizer once per communication interval. Under a fixed local horizon, every outer update
follows the same configured number of inner steps. Once the horizon varies,
successive updates no longer share this common outer clock. A longer interval
accumulate more inner learning rate, change the scale of the averaged
pseudo-gradient, and allow the incoming momentum state to persist across more
inner optimization.

Prior work motivates treating the update after local computation as a distinct
optimization layer. SlowMo and SNOO study momentum on the outer optimization
clock, while FedNova and adaptive federated optimization account for the local
computation represented by an aggregated update
\citep{wang2020slowmo,kallusky2025snoo,wang2020fednova,
reddi2021adaptive}. AutoLoCo addresses the corresponding problem across
successive DiLoCo intervals. Workers use one shared horizon within each
interval, but that horizon may change between intervals. We therefore compare
the selected interval with one base token interval at the same scheduler
position and use this comparison throughout the outer correction.

\subsection{Scheduler-Aligned Interval Comparison}
\label{app:outer-correction-interval-comparison}

Let \(s_t\) denote the global inner-step index at the beginning of interval
\(t\). The token mapper provides \(H_t^{\mathrm{step}}\), the executable
horizon selected for this interval, and \(H_t^{\mathrm{ref}}\), the executable
horizon corresponding to one base token interval at the current token density.
Starting from \(s_t\), their accumulated inner learning rates are
\begin{equation}
\Lambda_t
=
\sum_{h=1}^{H_t^{\mathrm{step}}}\eta_{t,h}
,
\qquad
\Lambda_t^{\mathrm{base}}
=
\sum_{h=1}^{H_t^{\mathrm{ref}}}\eta_{t,h}
.
\label{eq:app-outer-correction-lr-masses}
\end{equation}
Here, \(\Lambda_t\) is the accumulated inner learning rate of interval \(t\),
and \(\Lambda_t^{\mathrm{base}}\) is the base-interval comparator evaluated
from the same scheduler position. The superscript \(\mathrm{base}\) refers to
the underlying DiLoCo recipe. It is distinct from
\(\Lambda_t^{\mathrm{ref}}\), which is the empirical accumulated
inner-learning-rate statistic stored in the accepted-horizon reference
profile.

For the detailed calculation, we write
\begin{equation}
\rho_t^{\mathrm{unc}}
=
\frac{
\Lambda_t
}{
\Lambda_t^{\mathrm{base}}+\epsilon
},
\qquad
\rho_t
=
\max
\left(
\rho_t^{\mathrm{unc}},
1
\right).
\label{eq:app-outer-correction-ratios}
\end{equation}
The value \(\rho_t^{\mathrm{unc}}\) is the interval ratio before the lower
bound is applied. The outer correction uses \(\rho_t\). An interval whose
accumulated inner learning rate does not exceed the base comparator therefore
uses the reference value \(\rho_t=1\).

Both scheduled sums become available once
\(H_t^{\mathrm{step}}\) and \(H_t^{\mathrm{ref}}\) have been determined.
Our implementation evaluates exact scheduler previews of these sums before
interval \(t\) begins and stores the resulting correction context for that
interval. After local execution, the completed \(\Lambda_t\) is recorded for
the adaptive-horizon reference profile. The completed measurement and the
preview cover the same scheduled steps for a full, non-truncated interval, but
they serve different parts of AutoLoCo. The preview configures the outer
correction, and the completed measurement supports the next horizon
decision.

\subsection{Pseudo-Gradient Normalization}
\label{app:outer-correction-normalization}

Workers first complete the selected local horizon and average their
pseudo-gradients into \(\bar{\Delta}_t\). We then apply
Equation~(\ref{eq:normalized-pseudo-gradient}) to this averaged quantity,
immediately before the outer optimizer step. Every worker therefore uses the
same \(\rho_t\) and the same normalized pseudo-gradient
\(\widetilde{\Delta}_t\).

The mathematical definition divides by \(\rho_t\). The implementation uses
\(\rho_t+\epsilon\) in the denominator for numerical protection. At
\(\rho_t=1\), normalization is the identity. As \(\rho_t\) increases, the
outer-optimizer input is reduced in proportion to the additional accumulated
inner learning rate represented by the selected interval. Pseudo-gradient normalization is enabled independently of the momentum and
outer-learning-rate adjustments.

\subsection{Momentum Retention and Bounded Outer-Step Scaling}
\label{app:outer-correction-momentum-step}

Under the base recipe, one outer update retains a fraction
\(\mu_{\mathrm{base}}\) of the incoming momentum state. If interval \(t\)
represents \(\rho_t\) base intervals, repeated application of this retention
gives \(\mu_{\mathrm{base}}^{\rho_t}\). 
For integer \(\rho_t\), the expression inside the clipping operation equals
the retention obtained from repeated base-momentum updates. The exponent
extends the same relation to non-integer interval ratios. This is described in detail in Appendix~\ref{app:outer_correction_derivation}.
The bounds \(\mu_{\min}\) and \(\mu_{\max}\) constrain the amount of momentum
retained by one outer update.

The interval ratio also determines the outer-step scale:
\begin{equation}
\kappa_t
=
\min
\left(
\rho_t,
\kappa_{\max}
\right).
\label{eq:app-outer-step-scale}
\end{equation}
We choose the outer learning rate so that its momentum-adjusted scale is
\(\kappa_t\) times the base value:
\begin{equation}
\frac{
\eta_t^{\mathrm{out}}
}{
1-\mu_t
}
=
\kappa_t
\frac{
\eta_{\mathrm{base}}^{\mathrm{out}}
}{
1-\mu_{\mathrm{base}}
}.
\label{eq:app-outer-step-ratio}
\end{equation}
Solving this relation gives
\begin{equation}
\eta_t^{\mathrm{out}}
=
\eta_{\mathrm{base}}^{\mathrm{out}}
\,
\kappa_t
\,
\frac{
1-\mu_t
}{
1-\mu_{\mathrm{base}}
}.
\label{eq:app-adjusted-outer-learning-rate}
\end{equation}
The momentum exponent changes retention over the interval, while
\(\kappa_t\) limits the corresponding increase in the
momentum-adjusted outer-step scale.

At \(\rho_t=1\), the interval ratio recovers the base case. Provided that
\(\mu_{\mathrm{base}}\in[\mu_{\min},\mu_{\max}]\), we obtain
\[
\widetilde{\Delta}_t
=
\bar{\Delta}_t,
\qquad
\mu_t
=
\mu_{\mathrm{base}},
\qquad
\kappa_t
=
1,
\qquad
\eta_t^{\mathrm{out}}
=
\eta_{\mathrm{base}}^{\mathrm{out}}.
\]
For \(\rho_t>1\), the correction lowers momentum retention and increases the
momentum-adjusted outer-step scale only up to \(\kappa_{\max}\). We update the
momentum coefficient without rescaling the stored momentum state.

\subsection{The AutoLoCo Algorithm}
\label{app:algorithm}

\begin{algorithm}[!ht]
\caption{AutoLoCo outer interval}
\label{alg:autoloco-outer-interval}
\begin{algorithmic}[1]
\Require synchronized parameters \(\theta_t\), incoming momentum state
\(v_{t-1}\), selected token-equivalent horizon
\(H_t^{\mathrm{tok}}\), accepted-horizon reference profile and
controller state, \(M_{\mathrm{base}},\hat n_{t-1},H_{\mathrm{base}},q_H,\beta,\epsilon\),
and inner learning rates \(\eta_{t,h}\), \(\eta_{\mathrm{base}}^{\mathrm{out}},
\mu_{\mathrm{base}},\mu_{\min},\mu_{\max},\kappa_{\max}\)

\Statex \textbf{Horizon and outer-correction setup}
\State Compute
\(H_t^{\mathrm{tok}},\,
H_t^{\mathrm{step}},\,
H_t^{\mathrm{ref}},\,
\rho_t,\,
\mu_t,\,
\kappa_t,\,
\eta_t^{\mathrm{out}}\)

\Statex \textbf{Local inner loops}
\For{\(i\gets1\) to \(N\) \textbf{ in parallel}}
    \State
    \(\displaystyle
    \theta_{t,0}^{(i)}
    \gets
    \theta_t
    \)
    \For{\(h\gets1\) to \(H_t^{\mathrm{step}}\)}
        \State
        \(\displaystyle
        \theta_{t,h}^{(i)}
        \gets
        \operatorname{InnerOpt}
        \left(
        \theta_{t,h-1}^{(i)};
        \mathcal B_{t,h}^{(i)},
        \eta_{t,h}
        \right)
        \)
    \EndFor
    \State
    \(\displaystyle
    \Delta_t^{(i)}
    \gets
    \theta_t-
    \theta_{t,H_t^{\mathrm{step}}}^{(i)}
    \)
\EndFor

\Statex \textbf{Synchronization and corrected outer update}
\State
\(\displaystyle
\bar{\Delta}_t
\gets
\frac{1}{N}
\sum_{i=1}^{N}
\Delta_t^{(i)}
\)
\Comment{synchronize pseudo-gradients}
\State
\(\displaystyle
\widetilde{\Delta}_t
\gets
\frac{
\bar{\Delta}_t
}{
\rho_t
}
\)
\State
\(\displaystyle
v_t
\gets
\mu_t v_{t-1}
+
\widetilde{\Delta}_t
\)
\State
\(\displaystyle
d_t
\gets
\widetilde{\Delta}_t
+
\mu_t v_t
\)
\State
\(\displaystyle
\theta_{t+1}
\gets
\theta_t
-
\eta_t^{\mathrm{out}}d_t
\)

\Statex \textbf{Interval evaluation and next horizon}
\State
\(\displaystyle
D_t
\gets
\frac{1}{N}
\sum_{i=1}^{N}
\left\|
\Delta_t^{(i)}
\right\|_2^2,
\qquad
C_t
\gets
\frac{
N\left\|
\bar{\Delta}_t
\right\|_2^2
}{
D_t+\epsilon
}
\)
\State
\(\displaystyle
\Lambda_t
\gets
\sum_{h=1}^{H_t^{\mathrm{step}}}
\eta_{t,h},
\qquad
M_t
\gets
\sum_{h=1}^{H_t^{\mathrm{step}}}
n_{t,h}
\)
\State
\(\displaystyle
\bar n_t
\gets
\frac{M_t}{H_t^{\mathrm{step}}},
\qquad
\hat n_t
\gets
\beta\hat n_{t-1}
+
(1-\beta)\bar n_t
\)
\State Assess \(D_t\), \(C_t\), and \(\Lambda_t\). Determine the current training status.
\State Update the reference profile and select
\(H_{t+1}^{\mathrm{tok}}\)
\State \Return
\(\theta_{t+1}\), \(v_t\), and \(H_{t+1}^{\mathrm{tok}}\)
\end{algorithmic}
\end{algorithm}

\section{Theoretical Analysis of AutoLoCo}
\label{app:theoretical-analysis}

\subsection{Interpretation of Local-Drift Energy and Aggregation Coherence}
\label{app:dc_geometric_interpretation}

During \(H_t^{\mathrm{step}}\) local updates without communication,
workers develop different parameter deviations from their shared
starting point, a phenomenon known as worker drift.
We want to characterize the relationship between this drift and the
synchronization interval.
To solve this problem, we relate the local-drift energy \(D_t\) and aggregation
coherence \(C_t\) to disagreement among the local endpoints of a
completed interval.
Suppose all \(N\) workers start interval \(t\) from \(\theta_t\) and complete
\(H_t^{\mathrm{step}}\) local updates. Their pseudo-gradients and exact
arithmetic mean are
\[
\Delta_t^{(i)}
=\theta_t-\theta_{t,H_t^{\mathrm{step}}}^{(i)},
\qquad
\bar{\Delta}_t=\frac{1}{N}\sum_{i=1}^{N}\Delta_t^{(i)}.
\]
For the numerical constant \(\epsilon>0\), the diagnostics are
\[
D_t=\frac{1}{N}\sum_{i=1}^{N}
\bigl\lVert\Delta_t^{(i)}\bigr\rVert_2^2,
\qquad
C_t=\frac{N\lVert\bar{\Delta}_t\rVert_2^2}{D_t+\epsilon}.
\]
The derivation applies to any inner optimizer and allows
\(H_t^{\mathrm{step}}\) to vary across intervals.

The pseudo-gradient sign convention gives the arithmetic mean of the
local endpoints as
\begin{equation}
\begin{aligned}
\frac{1}{N}\sum_{i=1}^{N}\theta_{t,H_t^{\mathrm{step}}}^{(i)}
&=\frac{1}{N}\sum_{i=1}^{N}
\bigl(\theta_t-\Delta_t^{(i)}\bigr)\\
&=\theta_t-\bar{\Delta}_t.
\end{aligned}
\label{eq:dc_endpoint_mean}
\end{equation}
This arithmetic mean need not equal the synchronized parameters
\(\theta_{t+1}\) produced by the Nesterov outer update.
Centering each local endpoint around this mean gives
\begin{equation}
\begin{aligned}
\theta_{t,H_t^{\mathrm{step}}}^{(i)}
-\bigl(\theta_t-\bar{\Delta}_t\bigr)
&=\bigl(\theta_t-\Delta_t^{(i)}\bigr)
-\bigl(\theta_t-\bar{\Delta}_t\bigr)\\
&=\bar{\Delta}_t-\Delta_t^{(i)}.
\end{aligned}
\label{eq:dc_centered_endpoints}
\end{equation}
Taking squared Euclidean norms and averaging therefore reduces endpoint
disagreement to the mean squared deviation of the pseudo-gradients.
Expanding this quantity yields
\begin{equation}
\begin{aligned}
&\frac{1}{N}\sum_{i=1}^{N}
\bigl\lVert\Delta_t^{(i)}-\bar{\Delta}_t\bigr\rVert_2^2
\\
&\qquad=\frac{1}{N}\sum_{i=1}^{N}
\left(
\bigl\lVert\Delta_t^{(i)}\bigr\rVert_2^2
-2\bigl\langle\Delta_t^{(i)},\bar{\Delta}_t\bigr\rangle
+\lVert\bar{\Delta}_t\rVert_2^2
\right)
\\
&\qquad=D_t
-2\left\langle
\frac{1}{N}\sum_{i=1}^{N}\Delta_t^{(i)},
\bar{\Delta}_t
\right\rangle
+\lVert\bar{\Delta}_t\rVert_2^2
\\
&\qquad=D_t-\lVert\bar{\Delta}_t\rVert_2^2.
\end{aligned}
\label{eq:dc_variance_expansion}
\end{equation}

Since \(D_t+\epsilon>0\), the definition of \(C_t\) gives
\begin{equation}
\lVert\bar{\Delta}_t\rVert_2^2
=\frac{C_t(D_t+\epsilon)}{N}.
\label{eq:dc_coherent_component}
\end{equation}
Substituting into Equation~(\ref{eq:dc_variance_expansion}) expresses
the mean squared distance of the local endpoints from their arithmetic
mean in terms of \(D_t\) and \(C_t\):
\begin{equation}
\begin{aligned}
&\frac{1}{N}\sum_{i=1}^{N}
\left\lVert
\theta_{t,H_t^{\mathrm{step}}}^{(i)}
-\bigl(\theta_t-\bar{\Delta}_t\bigr)
\right\rVert_2^2
\\
&\qquad=\frac{1}{N}\sum_{i=1}^{N}
\bigl\lVert\Delta_t^{(i)}-\bar{\Delta}_t\bigr\rVert_2^2
\\
&\qquad=D_t-\frac{C_t(D_t+\epsilon)}{N}.
\end{aligned}
\label{eq:dc_endpoint_disagreement}
\end{equation}
This shows that \(D_t\) and \(C_t\) jointly provide an exact characterization of the disagreement among worker endpoints accumulated during a local synchronization interval. The endpoint disagreement is \(D_t-C_t(D_t+\epsilon)/N\).

The squared norm is convex, so
\begin{equation}
\lVert\bar{\Delta}_t\rVert_2^2
\le\frac{1}{N}\sum_{i=1}^{N}
\bigl\lVert\Delta_t^{(i)}\bigr\rVert_2^2=D_t.
\label{eq:dc_mean_norm_bound}
\end{equation}
Consequently, for finite pseudo-gradients,
\begin{equation}
\begin{aligned}
0&\le C_t\le\frac{ND_t}{D_t+\epsilon}<N,\\
0&\le D_t-\frac{C_t(D_t+\epsilon)}{N}\le D_t.
\end{aligned}
\label{eq:dc_ranges}
\end{equation}

So, \(C_t\) is a bounded measure of coherence, while the endpoint disagreement determined by \(D_t\) and \(C_t\) is nonnegative and cannot exceed the total drift energy \(D_t\).

If \(D_t=0\), then \(\Delta_t^{(i)}=0\) for every worker, so \(C_t=0\)
and all local endpoints coincide with \(\theta_t\).
More generally, if all worker pseudo-gradients are identical, the local
endpoints coincide and
\begin{equation}
C_t=\frac{ND_t}{D_t+\epsilon},
\label{eq:dc_identical_updates}
\end{equation}
attaining the upper bound for the given \(D_t\).

The Interpretation is, 
\(D_t\) and \(C_t\) distinguish large but coherent local motion
from motion with substantial disagreement across workers.
Local-drift energy \(D_t\) measures the mean squared displacement of
the local endpoints from the synchronized parameters \(\theta_t\).
Aggregation coherence \(C_t\) quantifies the squared norm retained
after averaging pseudo-gradients, relative to \(D_t+\epsilon\)
and scaled by \(N\).
For fixed \(N\) and \(D_t\), a larger \(C_t\) therefore implies
smaller endpoint disagreement.

\subsection{Learning-Rate Mass and Scheduler-Aligned Interval Ratios}
\label{app:lr_mass_interval_ratio}

Changing the number of local inner steps changes the local computation
represented by each outer update and can alter the scale of the averaged
pseudo-gradient. Fixed outer coefficients do not explicitly account for
these changes. AutoLoCo therefore uses outer correction to adjust
momentum retention and the outer learning rate for the selected interval.
This correction requires a measure of the interval's scale relative to
the base recipe. We measure this scale using the ratio of accumulated
inner learning rates, or learning-rate masses. We then examine when this
ratio also describes the relative scale of local displacements.

The token mapper provides \(H_t^{\mathrm{step}}\) and
\(H_t^{\mathrm{ref}}\), the selected executable horizon and the executable
horizon corresponding to one base token interval at the current token
density. Both horizons start at the same position in the inner
learning-rate schedule. Their learning-rate masses are
\begin{equation}
\Lambda_t
=\sum_{h=1}^{H_t^{\mathrm{step}}}\eta_{t,h},
\qquad
\Lambda_t^{\mathrm{base}}
=\sum_{h=1}^{H_t^{\mathrm{ref}}}\eta_{t,h}.
\label{eq:lr_ratio_masses}
\end{equation}
The comparator \(\Lambda_t^{\mathrm{base}}\) is evaluated at the current
scheduler position. It is not the historical
\(\Lambda_t^{\mathrm{ref}}\) stored in the accepted-horizon reference
profile. This choice measures the effect of the selected horizon relative
to the current base interval, rather than comparing different training
stages. For the displacement analysis below, assume that the inner
learning rates over the compared ranges are positive. The protected ratio
defined later also accommodates nonnegative learning rates.

\paragraph{Learning-rate mass and local displacement.}
Using the pseudo-gradient sign convention, we telescope the local updates
to obtain
\begin{equation}
\begin{aligned}
\Delta_t^{(i)}
&=\theta_t-\theta_{t,H_t^{\mathrm{step}}}^{(i)}\\
&=\sum_{h=1}^{H_t^{\mathrm{step}}}
\bigl(\theta_{t,h-1}^{(i)}-\theta_{t,h}^{(i)}\bigr)\\
&=\sum_{h=1}^{H_t^{\mathrm{step}}}
\eta_{t,h}
\frac{\theta_{t,h-1}^{(i)}-\theta_{t,h}^{(i)}}{\eta_{t,h}}.
\end{aligned}
\label{eq:lr_ratio_displacement_sum}
\end{equation}
The quotient in the last line is the effective parameter displacement
per unit inner learning rate. It includes the effects of the inner
optimizer.

Dividing Equation~(\ref{eq:lr_ratio_displacement_sum}) by
\(\Lambda_t>0\) yields
\begin{equation}
\begin{aligned}
\frac{\Delta_t^{(i)}}{\Lambda_t}
&=\sum_{h=1}^{H_t^{\mathrm{step}}}
\frac{\eta_{t,h}}{\Lambda_t}
\frac{\theta_{t,h-1}^{(i)}-\theta_{t,h}^{(i)}}{\eta_{t,h}},\\
\sum_{h=1}^{H_t^{\mathrm{step}}}
\frac{\eta_{t,h}}{\Lambda_t}&=1.
\end{aligned}
\label{eq:lr_ratio_weighted_average}
\end{equation}
Here, \(\Lambda_t\) is the total learning-rate weight, and
\(\Delta_t^{(i)}/\Lambda_t\) is a weighted average of the effective local
updates. Then, interpreting local displacement
as proportional to learning-rate mass additionally requires the effective
updates to remain sufficiently close over the compared steps.

\paragraph{Comparison with the base interval.}
Consider \(H_t^{\mathrm{step}}>H_t^{\mathrm{ref}}\).
The point \(\theta_{t,H_t^{\mathrm{ref}}}^{(i)}\) is the base-horizon
prefix endpoint of the same local trajectory. It therefore shares the
initial parameters, local optimizer state, and minibatch prefix with the
selected interval, and no intermediate outer update is inserted. We have
\begin{equation}
\begin{aligned}
\Lambda_t-\Lambda_t^{\mathrm{base}}
&=\sum_{h=H_t^{\mathrm{ref}}+1}^{H_t^{\mathrm{step}}}
\eta_{t,h},\\
\Delta_t^{(i)}
-\bigl(\theta_t-\theta_{t,H_t^{\mathrm{ref}}}^{(i)}\bigr)
&=\sum_{h=H_t^{\mathrm{ref}}+1}^{H_t^{\mathrm{step}}}
\bigl(\theta_{t,h-1}^{(i)}-\theta_{t,h}^{(i)}\bigr).
\end{aligned}
\label{eq:lr_ratio_prefix_decomposition}
\end{equation}
Subtracting the base-prefix displacement scaled by
\(\Lambda_t/\Lambda_t^{\mathrm{base}}\) gives
\begin{equation}
\begin{aligned}
&\Delta_t^{(i)}
-\frac{\Lambda_t}{\Lambda_t^{\mathrm{base}}}
\bigl(\theta_t-\theta_{t,H_t^{\mathrm{ref}}}^{(i)}\bigr)
\\
&\quad=
\sum_{h=H_t^{\mathrm{ref}}+1}^{H_t^{\mathrm{step}}}
\left[
\theta_{t,h-1}^{(i)}-\theta_{t,h}^{(i)}
-\frac{\eta_{t,h}}{\Lambda_t^{\mathrm{base}}}
\bigl(\theta_t-\theta_{t,H_t^{\mathrm{ref}}}^{(i)}\bigr)
\right]
\\
&\quad=
\sum_{h=H_t^{\mathrm{ref}}+1}^{H_t^{\mathrm{step}}}
\eta_{t,h}
\left[
\frac{\theta_{t,h-1}^{(i)}-\theta_{t,h}^{(i)}}{\eta_{t,h}}
-\frac{\theta_t-\theta_{t,H_t^{\mathrm{ref}}}^{(i)}}
{\Lambda_t^{\mathrm{base}}}
\right].
\end{aligned}
\label{eq:lr_ratio_displacement_remainder}
\end{equation}
By the triangle inequality and positivity of the learning rates,
\begin{equation}
\begin{aligned}
&\left\lVert
\Delta_t^{(i)}
-\frac{\Lambda_t}{\Lambda_t^{\mathrm{base}}}
\bigl(\theta_t-\theta_{t,H_t^{\mathrm{ref}}}^{(i)}\bigr)
\right\rVert_2
\\
&\quad\le
\sum_{h=H_t^{\mathrm{ref}}+1}^{H_t^{\mathrm{step}}}
\eta_{t,h}
\left\lVert
\frac{\theta_{t,h-1}^{(i)}-\theta_{t,h}^{(i)}}{\eta_{t,h}}
-\frac{\theta_t-\theta_{t,H_t^{\mathrm{ref}}}^{(i)}}
{\Lambda_t^{\mathrm{base}}}
\right\rVert_2
\\
&\quad\le
\bigl(\Lambda_t-\Lambda_t^{\mathrm{base}}\bigr)
\max_{H_t^{\mathrm{ref}}<h\le H_t^{\mathrm{step}}}
\left\lVert
\frac{\theta_{t,h-1}^{(i)}-\theta_{t,h}^{(i)}}{\eta_{t,h}}
-\frac{\theta_t-\theta_{t,H_t^{\mathrm{ref}}}^{(i)}}
{\Lambda_t^{\mathrm{base}}}
\right\rVert_2.
\end{aligned}
\label{eq:lr_ratio_displacement_bound}
\end{equation}
Equation~(\ref{eq:lr_ratio_displacement_bound}) bounds the error in
representing the selected-interval displacement as the base-prefix
displacement scaled by \(\Lambda_t/\Lambda_t^{\mathrm{base}}\).
The bound is the additional learning-rate mass multiplied by the largest
deviation of the appended effective updates from the learning-rate-weighted
mean of the base prefix. When this product is small, the selected
interval's displacement is close to this scaled base-prefix displacement.
The relation is exact when every appended effective update equals the
base-prefix mean. So, the LR-mass ratio has a local-displacement
interpretation under the stated condition on effective-update variation.

For the outer correction, AutoLoCo uses the lower-bounded ratio
\(
\rho_t=\max\left(
\frac{\Lambda_t}{\Lambda_t^{\mathrm{base}}+\epsilon},1
\right).
\)
We next use \(\rho_t\) to coordinate momentum retention and bounded
outer-step scaling relative to the base recipe.

\subsection{Momentum Retention and Bounded Outer-Step Scaling}
\label{app:outer_correction_derivation}

The interval ratio relates the selected interval to the base recipe.
We use this ratio to adjust momentum retention and scale the resulting
outer direction. In the version without pseudo-gradient normalization,
the Nesterov outer update is
\begin{equation}
\begin{aligned}
v_t &= \mu_t v_{t-1}+\bar{\Delta}_t,\\
d_t &= \bar{\Delta}_t+\mu_t v_t,\\
u_t &= \theta_t-\theta_{t+1}
     =\eta_t^{\mathrm{out}}d_t.
\end{aligned}
\label{eq:oc_unnormalized_update}
\end{equation}

A fixed momentum coefficient discounts the incoming state once per
synchronization, regardless of the local computation between
synchronizations. To motivate the retention rule before clipping, we hold
the base unit of learning-rate mass fixed and consider only the
contribution of the incoming state. One base unit retains a fraction
\(\mu_{\mathrm{base}}\), where \(0<\mu_{\mathrm{base}}<1\).
We require retention to depend only on the accumulated exposure and to
compose multiplicatively when that exposure is partitioned. For a positive,
continuous retention rule, this makes log retention additive and therefore
linear in exposure. Calibration at one base unit gives log retention
\(\rho_t\log\mu_{\mathrm{base}}\) at \(\rho_t\) units.
The retained fraction is therefore
\begin{equation}
\exp\!\left(\rho_t\log\mu_{\mathrm{base}}\right)
=\mu_{\mathrm{base}}^{\rho_t}.
\label{eq:oc_exponential_retention}
\end{equation}
This gives the exponential term in the interval-adjusted momentum rule.

Changing momentum also changes the response to repeated pseudo-gradient
inputs. To characterize this effect, fix \(\mu_t\),
\(\eta_t^{\mathrm{out}}\), and \(\bar{\Delta}_t\) at their interval-\(t\)
values and repeatedly apply the same outer recurrence, with
\(0\leq\mu_t<1\). This is a fixed-input response analysis, not an
assumption that training inputs remain constant.
Subtracting \(\bar{\Delta}_t/(1-\mu_t)\) from the momentum recurrence gives
\begin{equation}
\begin{aligned}
v_t-\frac{\bar{\Delta}_t}{1-\mu_t}
&=\mu_t v_{t-1}+\bar{\Delta}_t
  -\frac{\bar{\Delta}_t}{1-\mu_t}\\
&=\mu_t\left(
v_{t-1}-\frac{\bar{\Delta}_t}{1-\mu_t}
\right).
\end{aligned}
\label{eq:oc_momentum_fixed_point}
\end{equation}
Each repetition contracts the distance to
\(\bar{\Delta}_t/(1-\mu_t)\) by the factor \(\mu_t\), so the momentum
converges to this fixed point. At the fixed point, the direction and
update become
\begin{equation}
\begin{aligned}
d_t
&=\bar{\Delta}_t+\mu_t\frac{\bar{\Delta}_t}{1-\mu_t}
 =\frac{\bar{\Delta}_t}{1-\mu_t},\\
u_t
&=\eta_t^{\mathrm{out}}d_t
 =\frac{\eta_t^{\mathrm{out}}}{1-\mu_t}\bar{\Delta}_t.
\end{aligned}
\label{eq:oc_steady_response}
\end{equation}
The outer learning rate and momentum jointly determine the
fixed-input response coefficient
\(\eta_t^{\mathrm{out}}/(1-\mu_t)\).
To set this coefficient to \(\kappa_t\) times its base value, we require
\begin{equation}
\frac{\eta_t^{\mathrm{out}}}{1-\mu_t}
=\kappa_t
\frac{\eta_{\mathrm{base}}^{\mathrm{out}}}
     {1-\mu_{\mathrm{base}}}.
\label{eq:oc_scale_matching}
\end{equation}
Solving for the outer learning rate yields
\begin{equation}
\eta_t^{\mathrm{out}}
=\eta_{\mathrm{base}}^{\mathrm{out}}\kappa_t
\frac{1-\mu_t}{1-\mu_{\mathrm{base}}}.
\label{eq:oc_adjusted_outer_lr}
\end{equation}
The factor \((1-\mu_t)/(1-\mu_{\mathrm{base}})\) compensates for the effect
of momentum on the fixed-input response, while \(\kappa_t\) specifies the
desired relative scale. The choice
\(\kappa_t=\min(\rho_t,\kappa_{\max})\) is a bounded-scaling policy.

\section{Relation Between AutoLoCo and DiLoCo}
\label{app:autoloco-diloco-relation}

\subsection{DiLoCo Preliminaries}
\label{sec:diloco-preliminaries}

AutoLoCo preserves DiLoCo's inner and outer training loop, so we begin with one
outer interval. At the start of interval \(t\), all \(N\) workers share the
parameters \(\theta_t\). Worker \(i\) updates its local copy for
\(H_t^{\mathrm{step}}\) steps and reaches a local endpoint.
Equation~(\ref{eq:diloco-local-trajectory}) follows worker \(i\) from the
shared model to this endpoint and defines the resulting pseudo-gradient:
\begin{equation}
\begin{aligned}
\theta_{t,0}^{(i)}
&=
\theta_t,
\\
\theta_{t,h}^{(i)}
&=
\operatorname{InnerOpt}
\left(
\theta_{t,h-1}^{(i)};
\mathcal B_{t,h}^{(i)},
\eta_{t,h}
\right),
\qquad
h=1,\ldots,H_t^{\mathrm{step}},
\\
\Delta_t^{(i)}
&=
\theta_t-
\theta_{t,H_t^{\mathrm{step}}}^{(i)}.
\end{aligned}
\label{eq:diloco-local-trajectory}
\end{equation}
Here, \(h\) indexes the local optimizer steps. At step \(h\), worker \(i\)
processes minibatch \(\mathcal B_{t,h}^{(i)}\) with inner learning rate
\(\eta_{t,h}\). The horizon \(H_t^{\mathrm{step}}\) is the number of local
updates performed before the next synchronization. We use AdamW as
\(\operatorname{InnerOpt}\), following DiLoCo.

After the local trajectories end, the workers synchronize and average their
pseudo-gradients. The outer optimizer uses this average to produce the next
synchronized model:
\begin{equation}
\bar{\Delta}_t
=
\frac{1}{N}
\sum_{i=1}^{N}
\Delta_t^{(i)},
\qquad
\theta_{t+1}
=
\operatorname{OuterOpt}
\left(
\theta_t,
\bar{\Delta}_t
\right).
\label{eq:diloco-outer-update}
\end{equation}
Equation~(\ref{eq:diloco-outer-update}) omits the outer momentum state and
scalar optimizer settings. Section~\ref{sec:outer-optimizer-correction} specifies the scalar optimizer
settings, and Equation~(\ref{eq:autoloco-corrected-nesterov}) gives the
explicit update with its momentum state. DiLoCo combines
many local AdamW steps with a Nesterov outer update. DiLoCo sets \(H_t^{\mathrm{step}}=H_{\mathrm{fix}}\) in every
interval and keeps the outer learning rate and momentum constant. AutoLoCo
retains Equations~(\ref{eq:diloco-local-trajectory})
and~(\ref{eq:diloco-outer-update}), but allows the executable horizon and
scalar outer optimizer settings to vary across intervals.

\subsection{DiLoCo as a Special Case}
\label{sec:fixed-horizon-special-case}

AutoLoCo recovers fixed-horizon DiLoCo when the selected horizon and outer
optimizer remain at their base settings. Suppose the controller always selects
\(H_{\mathrm{base}}\), and suppose the mapper operates at the reference token
density. It then returns
\begin{equation}
H_t^{\mathrm{tok}}
=
H_{\mathrm{base}},
\qquad
H_t^{\mathrm{step}}
=
H_t^{\mathrm{ref}}
=
H_{\mathrm{base}},
\qquad
\forall t.
\label{eq:fixed-horizon-mapping}
\end{equation}

The selected interval and the base comparator now cover the same part of the
inner learning-rate schedule. Their accumulated inner learning rates are
therefore equal:
\begin{equation}
\Lambda_t
=
\Lambda_t^{\mathrm{base}},
\qquad
\rho_t
=
1.
\label{eq:fixed-horizon-ratio}
\end{equation}

Assume that
\(\mu_{\mathrm{base}}\in[\mu_{\min},\mu_{\max}]\). Substituting
\(\rho_t=1\) into the outer correction gives
\begin{equation}
\widetilde{\Delta}_t
=
\bar{\Delta}_t,
\qquad
\mu_t
=
\mu_{\mathrm{base}},
\qquad
\kappa_t
=
1,
\qquad
\eta_t^{\mathrm{out}}
=
\eta_{\mathrm{base}}^{\mathrm{out}}.
\label{eq:fixed-horizon-correction}
\end{equation}

With the same incoming momentum state, the Nesterov update becomes
\begin{equation}
\begin{aligned}
v_t
&=
\mu_{\mathrm{base}}v_{t-1}
+
\bar{\Delta}_t,
\\
d_t
&=
\bar{\Delta}_t
+
\mu_{\mathrm{base}}v_t,
\\
\theta_{t+1}
&=
\theta_t
-
\eta_{\mathrm{base}}^{\mathrm{out}}d_t.
\end{aligned}
\label{eq:fixed-horizon-nesterov}
\end{equation}
This is the DiLoCo outer update under the pseudo-gradient sign
convention in Equation~(\ref{eq:diloco-local-trajectory}). Thus, AutoLoCo reduces exactly to
DiLoCo in the base setting.

\section{Experimental settings}
\label{app:experimental_settings}

\subsection{Training and evaluation protocol}
\label{app:experimental_data_budget}

We simulate 8 independent workers for the main C4 pre-training and SFT experiments.
Each local-training worker maintains its own model replica and AdamW state. We also evaluate SFT with 16 workers communicating over a physical network.
Table~\ref{tab:exp_optimizer_settings} summarizes the training settings.
Within each comparison, methods share the model initialization, data, and training budget.

\begin{table}[htbp]
\centering
\caption{Training settings. Batch sizes count sequences per optimizer step. For SFT, the paired values correspond to eight and sixteen workers. The C4 ablations use four workers and a global batch of 512, with the remaining training settings unchanged.}
\label{tab:exp_optimizer_settings}
\label{tab:exp_batch_settings}
\vspace{\baselineskip}
\begin{tabular*}{\linewidth}{@{\extracolsep{\fill}}lll@{}}
\toprule
Setting & C4 pre-training & UltraChat SFT \\
\midrule
Model & LLaMA (215M total) & Llama-3.1-8B Base \\
Initialization & Shared random initialization & Shared pretrained checkpoint \\
Training budget & 50,000 steps & One epoch after filtering \\
Workers & 4 / 8 & 8 / 16 \\
Effective batch per worker & 128 & 4 / 2 \\
Global batch & 512 / 1,024 & 32 \\
Sequence length & 1,024 & At most 2,048 \\
Inner optimizer & AdamW & AdamW \\
Peak inner learning rate & \(4\times10^{-4}\) & \(5\times10^{-6}\) \\
AdamW coefficients & \((0.9,0.95)\) & \((0.9,0.999)\) \\
Weight decay & 0.1 & 0 \\
Gradient-norm clipping & 1.0 & 1.0 \\
Learning-rate schedule & Cosine & Cosine \\
Warm-up & 1,000 steps & 3\% of steps \\
Compute precision & BF16 & BF16 \\
Random seed & 42 & 42 \\
\bottomrule
\end{tabular*}
\end{table}

We shard the English training split across workers and truncate or pad each document to 1,024 tokens without packing.
The main budget therefore covers 52.43B input token positions, including padding; the 4 worker ablations cover 26.21B.

For SFT, we update all parameters of Llama-3.1-8B on UltraChat-200k.
We filter invalid conversations and examples without assistant targets after truncation, then train for one pass through the retained train data.
All SFT comparisons consume the same shuffled global batches of 32, with the final incomplete batch dropped.
Conversations are dynamically padded without packing, and only assistant target tokens contribute to the loss.
The token mapper counts all non-padding input tokens, including the conversational context.

We evaluate the final synchronized model using token-averaged negative log-likelihood.
Our C4 evaluator is configured to use 4,096 validation documents.
For SFT, we use a subset of 1,000 valid test conversations and average loss over assistant targets.

\subsection{Baselines and AutoLoCo configuration}
\label{app:experimental_optimization}

DDP synchronizes gradients at every optimizer update.
DiLoCo uses local AdamW and a Nesterov outer optimizer, with \(H_{\mathrm{fix}}=500\) for C4 and \(H_{\mathrm{fix}}=100\) for SFT.
DiLoCo follows the configuration
reported in prior work.

For QSR~\citep{gu2024quadratic}, we follow the inner learning rate division at the start of an interval, square the ratio, and round down, with a minimum of 500 steps.
Warm-up intervals remain fixed at 500 steps.
Linear-Interval~\citep{spiridonoff2021communicationefficient} uses intervals of 10, 20, \(\ldots\), 990 steps, giving 100 scheduled synchronizations within the 50,000 step budget. This matches the same number of synchronization compared with DiLoCo.
Both baselines use local AdamW with direct model averaging and retain worker-local optimizer states.

\begin{table}[htbp]
\centering
\caption{AutoLoCo configuration. Shared settings apply to both tasks.}
\label{tab:exp_controller_settings}
\vspace{\baselineskip}
\begin{tabular*}{\linewidth}{@{\extracolsep{\fill}}lll@{}}
\toprule
Setting & C4 & UltraChat \\
\midrule
Base token horizon \(H_{\mathrm{base}}\) & 500 & 100 \\
Token-horizon bounds \([H_{\min},H_{\max}]\) & \([250,750]\) & \([50,150]\) \\
Horizon quantum \(q_H\) & 50 & 10 \\
Base outer LR \(\eta_{\mathrm{base}}^{\mathrm{out}}\) & 0.7 & 0.7 \\
Base outer momentum \(\mu_{\mathrm{base}}\) & 0.7 / 0.9 & 0.7 \\
\midrule
Shared setting & \multicolumn{2}{l}{Value} \\
\midrule
Token-density EMA coefficient \(\beta\) & \multicolumn{2}{l}{0.9} \\
Mapper / reference bootstrap intervals & \multicolumn{2}{l}{3 / 2} \\
Candidate / moderate / severe thresholds & \multicolumn{2}{l}{1.5 / 2.5 / 4.0} \\
Candidate LR-exposure multiplier & \multicolumn{2}{l}{1.15} \\
Moderate / severe reduction divisors & \multicolumn{2}{l}{1.20 / 1.50} \\
Momentum bounds \([\mu_{\min},\mu_{\max}]\) & \multicolumn{2}{l}{\([0.45,0.90]\)} \\
Maximum outer-step scale \(\kappa_{\max}\) & \multicolumn{2}{l}{1.6} \\
\bottomrule
\end{tabular*}
\end{table}

\subsection{UltraChat Fine-Tuning and Evaluation Configuration}
\label{app:ultrachat-protocol}

\paragraph{Training settings.}
All runs in Section~\ref{sec:ultrachat-sft} use Llama-3.1-8B, a global
batch of 32, and 6,490 optimizer steps. The 8 workers comparison uses
8 DDP replicas or 8 independent local workers. The larger
comparison uses 16 local workers. DiLoCo synchronizes every 100 local
steps and uses outer momentum 0.9. AutoLoCo includes pseudo-gradient
normalization, adaptive momentum
and learning-rate correction.

\paragraph{Downstream evaluation.}
IFEval~\citep{zhou2023ifeval} and IFBench~\citep{pyatkin2025ifbench}
report prompt-level strict and loose accuracy, respectively.
MMLU-Pro~\citep{wang2024mmlupro} uses 5-shot generation and reports
accuracy. BBH~\citep{suzgun2023bbh} uses 3-shot chain-of-thought
prompting and reports micro-averaged exact match of the generated
answers. HumanEval+~\citep{liu2023evalplus} reports plus pass@1 across
164 tasks, with one greedy generation per task and both the base and
additional tests required to pass. All downstream scores are expressed
as percentages, and differences in the main text are percentage points.

\subsection{Ablation Configuration}
\label{app:component-ablation}

The experiment in Section~\ref{sec:component-ablation} uses 4 workers
on C4, a global batch of 512, and 50,000 optimizer steps. All
configurations share the initialization, training data, and inner AdamW
recipe, with seed 42. DiLoCo synchronizes every $H=500$ local steps. For a controlled comparison, we set DiLoCo's outer momentum to 0.7, the same value used by AutoLoCo \(\mu_{\mathrm{base}}\), instead of the 0.9 adopted in the earlier experiments.
The adaptive-interval variant enables horizon selection and token mapping
while retaining the base outer optimizer. The intervals + momentum
variant additionally adjusts momentum retention and keeps the outer
learning rate at its base value. Full AutoLoCo also enables the bounded
outer-step adjustment.

\section{Additional Results of Pre-training}
\label{app:pretrain}

\subsection{Training Trajectories and Thresholds}
\label{app:pretrain-trajectories}
\label{app:pretrain-thresholds}

At 25\%, 50\%, 75\%, and 100\% of the budget, AutoLoCo's
loss minus DiLoCo's is $+0.0187$, $+0.0026$, $-0.0060$, and $-0.0161$,
respectively. Each value uses the 1,000 steps ending at that progress
point. Table~\ref{tab:pretrain-threshold-times} reports the first crossings
of training-loss thresholds under this aggregation. AutoLoCo reaches
2.85 at step 37,336, compared with 38,217 for DiLoCo.

\begin{table}[!htbp]
\centering
\caption{Control accounting over 73
\mbox{AutoLoCo} rounds.
Each block uses the maximum worker duration per round.
Scalar bytes are included in endpoint accounting.}
\label{tab:pretrain-control-overhead}
\setlength{\tabcolsep}{3pt}
\vspace{.5\baselineskip}
\begin{tabularx}{\linewidth}{
    @{}>{\raggedright\arraybackslash}Xr@{}
}
\toprule
Block & Cost per round \\
\midrule
Preparation + local drift statistics
& 1,126.07~ms \\
Aggregated gradient norm for $C_t$
& 111.81~ms \\
Outer preparation
& 100.70~ms \\
Post-optimizer scalar/control span
& 1,090.98~ms \\
\midrule
Total
& 2,429.55~ms \\
\midrule
Scalar contributions (worker payload)
& 256~B \\
Control accounting (all endpoints)
& 40.56~KiB \\
\bottomrule
\end{tabularx}
\end{table}

\begin{figure}[!htbp]
\centering
\includegraphics[width=0.46\linewidth]
    {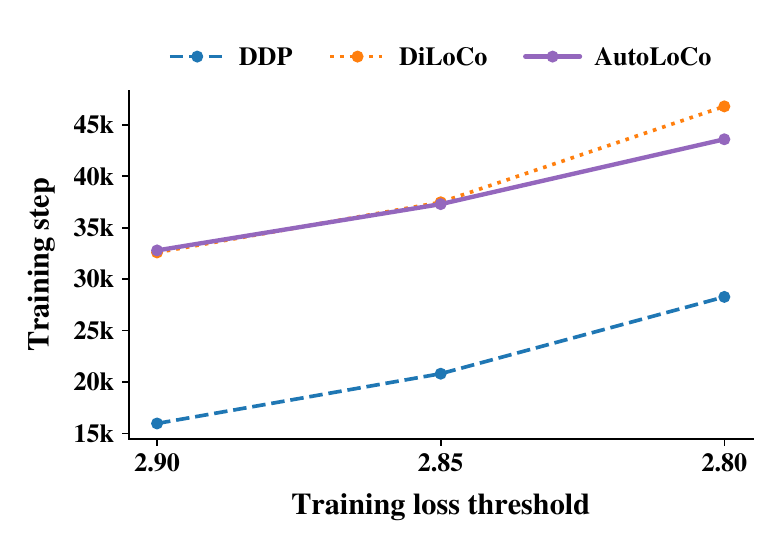}
\caption{Training steps at loss thresholds of
2.90, 2.85, and 2.80. It shows the number of training steps required to reach each loss value.}
\label{fig:pretrain-thresholds}
\end{figure}

\subsection{Executed Intervals and Outer Settings}
\label{app:pretrain-outer}
\label{app:pretrain-horizons}

AutoLoCo adopts five candidate horizons and executes the following
sequence, written as executable step horizon times number of intervals:
\(
500\times7, 550\times5, 600\times5, 650\times5, 
700\times5, 750\times45, 250\times1.
\)
The 73 intervals sum to 50,000 steps. The last 250 steps are the
remaining training budget. These runs exercise horizon growth and
operation at the upper bound.
Linear Interval executes horizons of 10, 20, $\ldots$, 990 steps,
followed by a 500-step remainder. QSR executes 41 intervals, with
median horizon 603.

\section{Additional Experiments of Low-precision communication}
\label{app:int8-communication}

Communicating with lower precision and reducing the frequency of synchronization 
are complementary ways to reduce communication overhead.
We combine AutoLoCo and INT8 in the 16-worker UltraChat experiment from
Section~\ref{sec:ultrachat-sft}.

\subsection{Low-precision communication Configuration}
\label{app:int8-configuration}

\paragraph{Codec and aggregation.}
Each tensor part is divided into contiguous blocks of up to 4,096
elements. For a nonzero block, the FP32 scale is its maximum absolute
value divided by 127. Elements are divided by this scale, rounded to
the nearest integer with ties to even, and clipped to $[-127,127]$.
A zero block uses scale 1. Scale bounds prevent underflow to zero
and overflow during reconstruction. Decoding multiplies the signed
INT8 values by the block scale in FP32. The codec uses no error
feedback. Worker contributions are decoded and averaged in FP32;
the mean is then encoded once for distribution. The worker responsible
for reducing a tensor part applies the same encoding and decoding
to its own contribution and receives the same decoded mean as the
other workers.

\paragraph{Training and controller.}
The INT8 runs retain the communication and evaluation recipe in
Appendices~\ref{app:experimental_data_budget} and~\ref{app:ultrachat-protocol}. DiLoCo uses 100 local steps
per interval. AutoLoCo uses pseudo-gradient
normalization, adaptive momentum and
learning-rate correction. Model computation remains in BF16, while
master parameters, AdamW states, the outer reference, and Nesterov
momentum remain in FP32. The controller measures $D_t$ from local
pseudo-gradients before quantization and computes $C_t$ using the
decoded aggregate. AutoLoCo uses the same horizon controller as in
the BF16 runs.

\subsection{Compatibility with Low-precision communication}
\label{sec:exp_int8}

With INT8 communication, AutoLoCo reduces total logical payload by
24.6\% relative to DiLoCo while improving test NLL from 0.9016 to
0.8772 (Table~\ref{tab:ultrachat_int8}).
For both methods, test NLL differs from the corresponding BF16 result
by less than 0.0001 (Table~\ref{tab:ultrachat_16worker_training}).
Combining AutoLoCo with INT8 reduces total logical payload by 62.3\%
relative to DiLoCo with BF16 communication.

\begin{table}[!ht]
    \centering
    \caption{INT8 pseudo-gradient communication with 16 workers on
    UltraChat. Optimizer settings follow Appendices~\ref{app:experimental_data_budget}--\ref{app:ultrachat-protocol}. INT8 applies to pseudo-gradient communication.}
    \label{tab:ultrachat_int8}
    \setlength{\tabcolsep}{3pt}
\vspace{\baselineskip}
    \begin{tabular*}{\linewidth}{@{\extracolsep{\fill}}lrrrrrr@{}}
        \toprule
        & & & \multicolumn{2}{c}{Logical payload (GB)}
        & \multicolumn{2}{c}{UltraChat test} \\
        \cmidrule(lr){4-5}
        \cmidrule(l){6-7}
        Method
        & \shortstack{Train loss $\downarrow$}
        & \shortstack{Syncs $\downarrow$}
        & Per sync
        & Total
        & NLL $\downarrow$
        & PPL $\downarrow$ \\
        \midrule
        DiLoCo (INT8)
        & 0.9004 & 65 & 8.038 & 522.477
        & 0.9016 & 2.4636 \\
        AutoLoCo (INT8)
        & 0.8761 & 49 & 8.038 & 393.867
        & 0.8772 & 2.4040 \\
        \bottomrule
    \end{tabular*}
\end{table}

\end{document}

%% file: math_commands.tex
\usepackage{amsmath,amsfonts,bm}

\def\eqref#1{equation~\ref{#1}}

\def\1{\bm{1}}

\DeclareMathAlphabet{\mathsfit}{\encodingdefault}{\sfdefault}{m}{sl}
\SetMathAlphabet{\mathsfit}{bold}{\encodingdefault}{\sfdefault}{bx}{n}













%% file: main.bbl
\begin{thebibliography}{44}
\providecommand{\natexlab}[1]{#1}
\providecommand{\url}[1]{\texttt{#1}}
\expandafter\ifx\csname urlstyle\endcsname\relax
  \providecommand{\doi}[1]{doi: #1}\else
  \providecommand{\doi}{doi: \begingroup \urlstyle{rm}\Url}\fi

\bibitem[Alistarh et~al.(2017)Alistarh, Grubic, Li, Tomioka, and
  Vojnovi{\'c}]{alistarh2017qsgd}
Dan Alistarh, Demjan Grubic, Jerry Li, Ryota Tomioka, and Milan Vojnovi{\'c}.
\newblock {QSGD}: Communication-efficient {SGD} via gradient quantization and
  encoding.
\newblock In \emph{Advances in Neural Information Processing Systems},
  volume~30, 2017.
\newblock URL
  \url{https://proceedings.neurips.cc/paper_files/paper/2017/hash/6c340f25839e6acdc73414517203f5f0-Abstract.html}.

\bibitem[Basu et~al.(2019)Basu, Data, Karakus, and
  Diggavi]{basu2019qsparselocalsgd}
Debraj Basu, Deepesh Data, Can Karakus, and Suhas Diggavi.
\newblock Qsparse-local-{SGD}: Distributed {SGD} with quantization,
  sparsification and local computations.
\newblock In \emph{Advances in Neural Information Processing Systems},
  volume~32, 2019.
\newblock URL
  \url{https://proceedings.neurips.cc/paper/2019/hash/d202ed5bcfa858c15a9f383c3e386ab2-Abstract.html}.

\bibitem[Charles et~al.(2025)Charles, Teston, Dery, Rush, Fallen, Garrett,
  Szlam, and Douillard]{charles2025dilocoscaling}
Zachary Charles, Gabriel Teston, Lucio Dery, John Rush, Nova Fallen, Zachary
  Garrett, Arthur~D. Szlam, and Arthur Douillard.
\newblock Communication-efficient language model training scales reliably and
  robustly: Scaling laws for {DiLoCo}.
\newblock In \emph{Advances in Neural Information Processing Systems}, 2025.
\newblock URL
  \url{https://papers.nips.cc/paper_files/paper/2025/hash/99acb4c087266e80b547aed79247266b-Abstract-Conference.html}.

\bibitem[Ding et~al.(2023)Ding, Chen, Xu, Qin, Zheng, Hu, Liu, Sun, and
  Zhou]{ding2023enhancing}
Ning Ding, Yulin Chen, Bokai Xu, Yujia Qin, Zhi Zheng, Shengding Hu, Zhiyuan
  Liu, Maosong Sun, and Bowen Zhou.
\newblock Enhancing chat language models by scaling high-quality instructional
  conversations.
\newblock \emph{arXiv preprint arXiv:2305.14233}, 2023.
\newblock URL \url{https://arxiv.org/abs/2305.14233}.

\bibitem[Douillard et~al.(2023)Douillard, Feng, Rusu, Chhaparia, Donchev,
  Kuncoro, Ranzato, Szlam, and Shen]{douillard2023diloco}
Arthur Douillard, Qixuan Feng, Andrei~A. Rusu, Rachita Chhaparia, Yani Donchev,
  Adhiguna Kuncoro, Marc'Aurelio Ranzato, Arthur Szlam, and Jiajun Shen.
\newblock {DiLoCo}: Distributed low-communication training of language models.
\newblock \emph{arXiv preprint arXiv:2311.08105}, 2023.
\newblock URL \url{https://arxiv.org/abs/2311.08105}.

\bibitem[Douillard et~al.(2025)Douillard, Donchev, Rush, Kale, Charles,
  Garrett, Teston, Lacey, McIlroy, Shen, Ram{\'e}, Szlam, Ranzato, and
  Barham]{douillard2025streaming}
Arthur Douillard, Yanislav Donchev, Keith Rush, Satyen Kale, Zachary Charles,
  Zachary Garrett, Gabriel Teston, Dave Lacey, Ross McIlroy, Jiajun Shen,
  Alexandre Ram{\'e}, Arthur Szlam, Marc'Aurelio Ranzato, and Paul Barham.
\newblock Streaming {DiLoCo} with overlapping communication: Towards a
  distributed free lunch.
\newblock \emph{arXiv preprint arXiv:2501.18512}, 2025.

\bibitem[Dubey et~al.(2024)Dubey, Jauhri, Pandey, Kadian, Al-Dahle, Letman,
  Mathur, Schelten, et~al.]{dubey2024llama3}
Abhimanyu Dubey, Abhinav Jauhri, Abhinav Pandey, Abhishek Kadian, Ahmad
  Al-Dahle, Aiesha Letman, Akhil Mathur, Alan Schelten, et~al.
\newblock The {Llama 3} herd of models.
\newblock \emph{arXiv preprint arXiv:2407.21783}, 2024.
\newblock URL \url{https://arxiv.org/abs/2407.21783v1}.

\bibitem[Gorbunov et~al.(2021)Gorbunov, Hanzely, and
  Richt{\'a}rik]{gorbunov2021localsgd}
Eduard Gorbunov, Filip Hanzely, and Peter Richt{\'a}rik.
\newblock Local {SGD}: Unified theory and new efficient methods.
\newblock In \emph{Proceedings of the 24th International Conference on
  Artificial Intelligence and Statistics}, pp.\  3556--3564. PMLR, 2021.
\newblock URL \url{https://proceedings.mlr.press/v130/gorbunov21a.html}.

\bibitem[Gu et~al.(2024)Gu, Lyu, Arora, Zhang, and Huang]{gu2024quadratic}
Xinran Gu, Kaifeng Lyu, Sanjeev Arora, Jingzhao Zhang, and Longbo Huang.
\newblock A quadratic synchronization rule for distributed deep learning.
\newblock In \emph{International Conference on Learning Representations}, 2024.
\newblock URL
  \url{https://proceedings.iclr.cc/paper_files/paper/2024/hash/128911cc894d57bcae78074a9551c132-Abstract-Conference.html}.

\bibitem[Haddadpour et~al.(2019)Haddadpour, Kamani, Mahdavi, and
  Cadambe]{haddadpour2019adaptive}
Farzin Haddadpour, Mohammad~Mahdi Kamani, Mehrdad Mahdavi, and Viveck Cadambe.
\newblock Local {SGD} with periodic averaging: Tighter analysis and adaptive
  synchronization.
\newblock In \emph{Advances in Neural Information Processing Systems},
  volume~32, 2019.
\newblock URL
  \url{https://proceedings.neurips.cc/paper_files/paper/2019/hash/c17028c9b6e0c5deaad29665d582284a-Abstract.html}.

\bibitem[Jaghouar et~al.(2024)Jaghouar, Ong, and
  Hagemann]{jaghouar2024opendiloco}
Sami Jaghouar, Jack~Min Ong, and Johannes Hagemann.
\newblock {OpenDiLoCo}: An open-source framework for globally distributed
  low-communication training.
\newblock \emph{arXiv preprint arXiv:2407.07852}, 2024.
\newblock URL \url{https://arxiv.org/abs/2407.07852}.

\bibitem[Jhunjhunwala et~al.(2023)Jhunjhunwala, Wang, and
  Joshi]{jhunjhunwala2023fedexp}
Divyansh Jhunjhunwala, Shiqiang Wang, and Gauri Joshi.
\newblock {FedExP}: Speeding up federated averaging via extrapolation.
\newblock In \emph{International Conference on Learning Representations}, 2023.
\newblock URL \url{https://openreview.net/forum?id=IPrzNbddXV}.

\bibitem[Jiang \& Agrawal(2020)Jiang and Agrawal]{jiang2020adaptive}
Peng Jiang and Gagan Agrawal.
\newblock Adaptive periodic averaging: A practical approach to reducing
  communication in distributed learning.
\newblock \emph{arXiv preprint arXiv:2007.06134}, 2020.
\newblock URL \url{https://arxiv.org/abs/2007.06134}.

\bibitem[Kallusky et~al.(2025)Kallusky, Rao, Nandavanam, and
  Shi]{kallusky2025snoo}
Dominik Kallusky, Vinay Rao, Vishal Nandavanam, and Hao-Jun~Michael Shi.
\newblock {SNOO}: Step-{K} nesterov outer optimizer---the surprising
  effectiveness of nesterov momentum applied to pseudo-gradients.
\newblock \emph{arXiv preprint arXiv:2510.15830}, 2025.
\newblock URL \url{https://arxiv.org/abs/2510.15830}.

\bibitem[Kamp et~al.(2018)Kamp, Adilova, Sicking, H{\"u}ger, Schlicht, Wirtz,
  and Wrobel]{kamp2018dynamicmodelaveraging}
Michael Kamp, Linara Adilova, Joachim Sicking, Fabian H{\"u}ger, Peter
  Schlicht, Tim Wirtz, and Stefan Wrobel.
\newblock Efficient decentralized deep learning by dynamic model averaging.
\newblock In \emph{Machine Learning and Knowledge Discovery in Databases}, pp.\
   393--409. Springer, 2018.
\newblock \doi{10.1007/978-3-030-10925-7_24}.
\newblock URL \url{https://doi.org/10.1007/978-3-030-10925-7_24}.

\bibitem[Karimireddy et~al.(2019)Karimireddy, Rebjock, Stich, and
  Jaggi]{karimireddy2019errorfeedback}
Sai~Praneeth Karimireddy, Quentin Rebjock, Sebastian~U. Stich, and Martin
  Jaggi.
\newblock Error feedback fixes {SignSGD} and other gradient compression
  schemes.
\newblock In \emph{Proceedings of the 36th International Conference on Machine
  Learning}, pp.\  3252--3261. PMLR, 2019.
\newblock URL \url{https://proceedings.mlr.press/v97/karimireddy19a.html}.

\bibitem[Karimireddy et~al.(2021)Karimireddy, Jaggi, Kale, Mohri, Reddi, Stich,
  and Suresh]{karimireddy2021mime}
Sai~Praneeth Karimireddy, Martin Jaggi, Satyen Kale, Mehryar Mohri, Sashank
  Reddi, Sebastian~U. Stich, and Ananda~Theertha Suresh.
\newblock Breaking the centralized barrier for cross-device federated learning.
\newblock In \emph{Advances in Neural Information Processing Systems},
  volume~34, 2021.
\newblock URL
  \url{https://proceedings.neurips.cc/paper_files/paper/2021/hash/f0e6be4ce76ccfa73c5a540d992d0756-Abstract.html}.

\bibitem[Khaled et~al.(2025)Khaled, Kale, Douillard, Jin, Fergus, and
  Zaheer]{khaled2025outeroptimizers}
Ahmed Khaled, Satyen Kale, Arthur Douillard, Chi Jin, Rob Fergus, and Manzil
  Zaheer.
\newblock Understanding outer optimizers in local {SGD}: Learning rates,
  momentum, and acceleration.
\newblock In \emph{Advances in Neural Information Processing Systems}, 2025.
\newblock URL \url{https://openreview.net/forum?id=2VX79YLT9s}.

\bibitem[Kim et~al.(2025)Kim, Li, Gandham, Baldonado, Gangidi, Balaji, Wang,
  and Akella]{kim2025halos}
Geon-Woo Kim, Junbo Li, Shashidhar Gandham, Omar Baldonado, Adithya Gangidi,
  Pavan Balaji, Zhangyang Wang, and Aditya Akella.
\newblock {HALoS}: Hierarchical asynchronous local {SGD} over slow networks for
  geo-distributed large language model training.
\newblock In \emph{Proceedings of the 42nd International Conference on Machine
  Learning}, pp.\  30548--30566. PMLR, 2025.
\newblock URL \url{https://proceedings.mlr.press/v267/kim25y.html}.

\bibitem[Li et~al.(2020)Li, Zhao, Varma, Salpekar, Noordhuis, Li, Paszke,
  Smith, Vaughan, Damania, and Chintala]{li2020pytorchdistributed}
Shen Li, Yanli Zhao, Rohan Varma, Omkar Salpekar, Pieter Noordhuis, Teng Li,
  Adam Paszke, Jeff Smith, Brian Vaughan, Pritam Damania, and Soumith Chintala.
\newblock {PyTorch} distributed: Experiences on accelerating data parallel
  training.
\newblock \emph{Proceedings of the VLDB Endowment}, 13\penalty0 (12):\penalty0
  3005--3018, 2020.
\newblock \doi{10.14778/3415478.3415530}.
\newblock URL \url{https://arxiv.org/abs/2006.15704}.

\bibitem[Lin et~al.(2020)Lin, Stich, Patel, and Jaggi]{lin2020postlocalsgd}
Tao Lin, Sebastian~U. Stich, Kumar~Kshitij Patel, and Martin Jaggi.
\newblock Don't use large mini-batches, use local {SGD}.
\newblock In \emph{International Conference on Learning Representations}, 2020.
\newblock URL \url{https://openreview.net/forum?id=B1eyO1BFPr}.

\bibitem[Lin et~al.(2018)Lin, Han, Mao, Wang, and
  Dally]{lin2018deepgradientcompression}
Yujun Lin, Song Han, Huizi Mao, Yu~Wang, and William~J. Dally.
\newblock Deep gradient compression: Reducing the communication bandwidth for
  distributed training.
\newblock In \emph{International Conference on Learning Representations}, 2018.
\newblock URL \url{https://openreview.net/forum?id=SkhQHMW0W}.

\bibitem[Liu et~al.(2023)Liu, Xia, Wang, and Zhang]{liu2023evalplus}
Jiawei Liu, Chunqiu~Steven Xia, Yuyao Wang, and Lingming Zhang.
\newblock Is your code generated by {ChatGPT} really correct? rigorous
  evaluation of large language models for code generation.
\newblock In \emph{Advances in Neural Information Processing Systems},
  volume~36, 2023.
\newblock URL
  \url{https://papers.neurips.cc/paper_files/paper/2023/hash/43e9d647ccd3e4b7b5baab53f0368686-Abstract-Conference.html}.

\bibitem[McMahan et~al.(2017)McMahan, Moore, Ramage, Hampson, and Ag{\"u}era~y
  Arcas]{mcmahan2017fedavg}
H.~Brendan McMahan, Eider Moore, Daniel Ramage, Seth Hampson, and Blaise
  Ag{\"u}era~y Arcas.
\newblock Communication-efficient learning of deep networks from decentralized
  data.
\newblock In \emph{Proceedings of the 20th International Conference on
  Artificial Intelligence and Statistics}, pp.\  1273--1282. PMLR, 2017.
\newblock URL \url{https://proceedings.mlr.press/v54/mcmahan17a.html}.

\bibitem[Pyatkin et~al.(2025)Pyatkin, Malik, Graf, Ivison, Huang, Dasigi,
  Lambert, and Hajishirzi]{pyatkin2025ifbench}
Valentina Pyatkin, Saumya Malik, Victoria Graf, Hamish Ivison, Shengyi Huang,
  Pradeep Dasigi, Nathan Lambert, and Hannaneh Hajishirzi.
\newblock Generalizing verifiable instruction following.
\newblock In \emph{Advances in Neural Information Processing Systems},
  volume~38, 2025.
\newblock URL \url{https://arxiv.org/abs/2507.02833}.

\bibitem[Raffel et~al.(2020)Raffel, Shazeer, Roberts, Lee, Narang, Matena,
  Zhou, Li, and Liu]{raffel2020exploring}
Colin Raffel, Noam Shazeer, Adam Roberts, Katherine Lee, Sharan Narang, Michael
  Matena, Yanqi Zhou, Wei Li, and Peter~J. Liu.
\newblock Exploring the limits of transfer learning with a unified text-to-text
  transformer.
\newblock \emph{Journal of Machine Learning Research}, 21\penalty0
  (140):\penalty0 1--67, 2020.
\newblock URL \url{https://jmlr.org/papers/v21/20-074.html}.

\bibitem[Reddi et~al.(2021)Reddi, Charles, Zaheer, Garrett, Rush,
  Kone{\v{c}}n{\'y}, Kumar, and McMahan]{reddi2021adaptive}
Sashank Reddi, Zachary Charles, Manzil Zaheer, Zachary Garrett, Keith Rush,
  Jakub Kone{\v{c}}n{\'y}, Sanjiv Kumar, and H.~Brendan McMahan.
\newblock Adaptive federated optimization.
\newblock In \emph{International Conference on Learning Representations}, 2021.
\newblock URL \url{https://openreview.net/forum?id=LkFG3lB13U5}.

\bibitem[Rothchild et~al.(2020)Rothchild, Panda, Ullah, Ivkin, Stoica,
  Braverman, Gonzalez, and Arora]{rothchild2020fetchsgd}
Daniel Rothchild, Ashwinee Panda, Enayat Ullah, Nikita Ivkin, Ion Stoica,
  Vladimir Braverman, Joseph~E. Gonzalez, and Raman Arora.
\newblock {FetchSGD}: Communication-efficient federated learning with
  sketching.
\newblock In \emph{Proceedings of the 37th International Conference on Machine
  Learning}, pp.\  8253--8265. PMLR, 2020.
\newblock URL \url{https://proceedings.mlr.press/v119/rothchild20a.html}.

\bibitem[Shen et~al.(2021)Shen, Cheng, Liu, and Xu]{shen2021stlsgd}
Shuheng Shen, Yifei Cheng, Jingchang Liu, and Linli Xu.
\newblock {STL-SGD}: Speeding up local {SGD} with stagewise communication
  period.
\newblock In \emph{Proceedings of the AAAI Conference on Artificial
  Intelligence}, volume~35, pp.\  9576--9584, 2021.
\newblock \doi{10.1609/aaai.v35i11.17153}.
\newblock URL \url{https://ojs.aaai.org/index.php/AAAI/article/view/17153}.

\bibitem[Spiridonoff et~al.(2021)Spiridonoff, Olshevsky, and
  Paschalidis]{spiridonoff2021communicationefficient}
Artin Spiridonoff, Alex Olshevsky, and Yannis Paschalidis.
\newblock Communication-efficient {SGD}: From local {SGD} to one-shot
  averaging.
\newblock In \emph{Advances in Neural Information Processing Systems},
  volume~34, 2021.
\newblock URL
  \url{https://proceedings.neurips.cc/paper/2021/hash/cc06a6150b92e17dd3076a0f0f9d2af4-Abstract.html}.

\bibitem[Stich(2019)]{stich2019localsgd}
Sebastian~U. Stich.
\newblock Local {SGD} converges fast and communicates little.
\newblock In \emph{International Conference on Learning Representations}, 2019.
\newblock URL \url{https://openreview.net/forum?id=S1g2JnRcFX}.

\bibitem[Suzgun et~al.(2023)Suzgun, Scales, Sch{\"a}rli, Gehrmann, Tay, Chung,
  Chowdhery, Le, Chi, Zhou, and Wei]{suzgun2023bbh}
Mirac Suzgun, Nathan Scales, Nathanael Sch{\"a}rli, Sebastian Gehrmann, Yi~Tay,
  Hyung~Won Chung, Aakanksha Chowdhery, Quoc Le, Ed~Chi, Denny Zhou, and Jason
  Wei.
\newblock Challenging {BIG-Bench} tasks and whether chain-of-thought can solve
  them.
\newblock In \emph{Findings of the Association for Computational Linguistics:
  ACL 2023}, pp.\  13003--13051. Association for Computational Linguistics,
  2023.
\newblock \doi{10.18653/v1/2023.findings-acl.824}.
\newblock URL \url{https://aclanthology.org/2023.findings-acl.824/}.

\bibitem[Tang et~al.(2021)Tang, Gan, Awan, Rajbhandari, Li, Lian, Liu, Zhang,
  and He]{tang2021onebitadam}
Hanlin Tang, Shaoduo Gan, Ammar~Ahmad Awan, Samyam Rajbhandari, Conglong Li,
  Xiangru Lian, Ji~Liu, Ce~Zhang, and Yuxiong He.
\newblock 1-bit {Adam}: Communication efficient large-scale training with
  {Adam}'s convergence speed.
\newblock In \emph{Proceedings of the 38th International Conference on Machine
  Learning}, pp.\  10118--10129. PMLR, 2021.
\newblock URL \url{https://proceedings.mlr.press/v139/tang21a.html}.

\bibitem[Th{\'e}rien et~al.(2026)Th{\'e}rien, Huang, Rish, and
  Belilovsky]{therien2026muloco}
Benjamin Th{\'e}rien, Xiaolong Huang, Irina Rish, and Eugene Belilovsky.
\newblock {MuLoCo}: Muon is a practical inner optimizer for {DiLoCo}.
\newblock In \emph{Proceedings of the 43rd International Conference on Machine
  Learning}, 2026.
\newblock URL \url{https://icml.cc/virtual/2026/poster/64378}.

\bibitem[Touvron et~al.(2023)Touvron, Lavril, Izacard, Martinet, Lachaux,
  Lacroix, Rozi{\`e}re, Goyal, Hambro, Azhar, Rodriguez, Joulin, Grave, and
  Lample]{touvron2023llama}
Hugo Touvron, Thibaut Lavril, Gautier Izacard, Xavier Martinet, Marie-Anne
  Lachaux, Timoth{\'e}e Lacroix, Baptiste Rozi{\`e}re, Naman Goyal, Eric
  Hambro, Faisal Azhar, Aurelien Rodriguez, Armand Joulin, Edouard Grave, and
  Guillaume Lample.
\newblock {LLaMA}: Open and efficient foundation language models.
\newblock \emph{arXiv preprint arXiv:2302.13971}, 2023.
\newblock URL \url{https://arxiv.org/abs/2302.13971}.

\bibitem[Vogels et~al.(2019)Vogels, Karimireddy, and Jaggi]{vogels2019powersgd}
Thijs Vogels, Sai~Praneeth Karimireddy, and Martin Jaggi.
\newblock {PowerSGD}: Practical low-rank gradient compression for distributed
  optimization.
\newblock In \emph{Advances in Neural Information Processing Systems},
  volume~32, 2019.
\newblock URL
  \url{https://proceedings.neurips.cc/paper_files/paper/2019/hash/d9fbed9da256e344c1fa46bb46c34c5f-Abstract.html}.

\bibitem[Wang \& Joshi(2019)Wang and Joshi]{wang2019adacomm}
Jianyu Wang and Gauri Joshi.
\newblock Adaptive communication strategies to achieve the best error-runtime
  trade-off in local-update {SGD}.
\newblock In \emph{Proceedings of Machine Learning and Systems}, volume~1,
  2019.
\newblock URL
  \url{https://proceedings.mlsys.org/paper_files/paper/2019/hash/4a0151b47bd93c5de2a0b57831981a0d-Abstract.html}.

\bibitem[Wang \& Joshi(2021)Wang and Joshi]{wang2021cooperativesgd}
Jianyu Wang and Gauri Joshi.
\newblock Cooperative {SGD}: A unified framework for the design and analysis of
  local-update {SGD} algorithms.
\newblock \emph{Journal of Machine Learning Research}, 22\penalty0
  (213):\penalty0 1--50, 2021.
\newblock URL \url{https://www.jmlr.org/papers/v22/20-147.html}.

\bibitem[Wang et~al.(2020{\natexlab{a}})Wang, Liu, Liang, Joshi, and
  Poor]{wang2020fednova}
Jianyu Wang, Qinghua Liu, Hao Liang, Gauri Joshi, and H.~Vincent Poor.
\newblock Tackling the objective inconsistency problem in heterogeneous
  federated optimization.
\newblock In \emph{Advances in Neural Information Processing Systems},
  volume~33, 2020{\natexlab{a}}.
\newblock URL
  \url{https://proceedings.neurips.cc/paper/2020/hash/564127c03caab942e503ee6f810f54fd-Abstract.html}.

\bibitem[Wang et~al.(2020{\natexlab{b}})Wang, Tantia, Ballas, and
  Rabbat]{wang2020slowmo}
Jianyu Wang, Vinayak Tantia, Nicolas Ballas, and Michael Rabbat.
\newblock {SlowMo}: Improving communication-efficient distributed {SGD} with
  slow momentum.
\newblock In \emph{International Conference on Learning Representations},
  2020{\natexlab{b}}.
\newblock URL \url{https://openreview.net/forum?id=SkxJ8REYPH}.

\bibitem[Wang et~al.(2023)Wang, Lu, Yuan, Chen, Liang, De~Sa, R{\'e}, and
  Zhang]{wang2023cocktailsgd}
Jue Wang, Yucheng Lu, Binhang Yuan, Beidi Chen, Percy Liang, Christopher De~Sa,
  Christopher R{\'e}, and Ce~Zhang.
\newblock {CocktailSGD}: Fine-tuning foundation models over 500 mbps networks.
\newblock In \emph{Proceedings of the 40th International Conference on Machine
  Learning}, pp.\  36058--36076. PMLR, 2023.
\newblock URL \url{https://proceedings.mlr.press/v202/wang23t.html}.

\bibitem[Wang et~al.(2024{\natexlab{a}})Wang, Ma, Zhang, Ni, Chandra, Guo, Ren,
  Arulraj, He, Jiang, Li, Ku, Wang, Zhuang, Fan, Yue, and
  Chen]{wang2024mmlupro}
Yubo Wang, Xueguang Ma, Ge~Zhang, Yuansheng Ni, Abhranil Chandra, Shiguang Guo,
  Weiming Ren, Aaran Arulraj, Xuan He, Ziyan Jiang, Tianle Li, Max Ku, Kai
  Wang, Alex Zhuang, Rongqi Fan, Xiang Yue, and Wenhu Chen.
\newblock {MMLU-Pro}: A more robust and challenging multi-task language
  understanding benchmark.
\newblock In \emph{Advances in Neural Information Processing Systems},
  volume~37, 2024{\natexlab{a}}.
\newblock URL
  \url{https://papers.nips.cc/paper_files/paper/2024/hash/ad236edc564f3e3156e1b2feafb99a24-Abstract-Datasets_and_Benchmarks_Track.html}.

\bibitem[Wang et~al.(2024{\natexlab{b}})Wang, Wang, Lu, and
  Chen]{wang2024fadas}
Yujia Wang, Shiqiang Wang, Songtao Lu, and Jinghui Chen.
\newblock {FADAS}: Towards federated adaptive asynchronous optimization.
\newblock In \emph{Proceedings of the 41st International Conference on Machine
  Learning}, pp.\  51701--51733. PMLR, 2024{\natexlab{b}}.
\newblock URL \url{https://proceedings.mlr.press/v235/wang24bv.html}.

\bibitem[Zhou et~al.(2023)Zhou, Lu, Mishra, Brahma, Basu, Luan, Zhou, and
  Hou]{zhou2023ifeval}
Jeffrey Zhou, Tianjian Lu, Swaroop Mishra, Siddhartha Brahma, Sujoy Basu,
  Yi~Luan, Denny Zhou, and Le~Hou.
\newblock Instruction-following evaluation for large language models.
\newblock \emph{arXiv preprint arXiv:2311.07911}, 2023.
\newblock URL \url{https://arxiv.org/abs/2311.07911}.

\end{thebibliography}
